\documentclass[sn-mathphys,Numbered]{sn-jnl}

\usepackage{graphicx}%
\usepackage{multirow}%
\usepackage{amsmath,amssymb,amsfonts}%
\usepackage{amsthm}%
\usepackage{mathrsfs}%
\usepackage[title]{appendix}%
\usepackage{xcolor}%
\usepackage{textcomp}%
\usepackage{manyfoot}%
\usepackage{booktabs}%
\usepackage{algorithm}%
\usepackage{algorithmicx}%
\usepackage{algpseudocode}%
\usepackage{listings}%
\usepackage{xcolor}

\theoremstyle{thmstyleone}%
\theoremstyle{thmstyletwo}%

\theoremstyle{thmstylethree}%

\begin{document}

\title[Article Title]{Applicability of Large Language Models and Generative Models for Legal Case Judgement Summarization}

\author*[1]{\fnm{Aniket} \sur{Deroy}}\email{roydanik18@kgpian.iitkgp.ac.in}

\author[2]{\fnm{Kripabandhu} \sur{Ghosh}}

\author[1]{\fnm{Saptarshi} \sur{Ghosh}}


\affil[1]{\orgdiv{Computer Science and Engineering}, \orgname{IIT Kharagpur}, \orgaddress{ \city{Kharagpur}, \postcode{721302}, \state{West Bengal}, \country{India}}}

\affil[2]{\orgdiv{Computational and Data Sciences}, \orgname{IISER Kolkata}, \orgaddress{\city{Kolkata}, \postcode{741246}, \state{West Bengal}, \country{India}}}









\abstract{
Automatic summarization of legal case judgements, which are known to be long and complex, has traditionally been tried via extractive summarization models. In recent years, generative models including abstractive summarization models and Large language models (LLMs) have gained huge popularity. In this paper, we explore the applicability of such models for legal case judgement summarization. 
We applied various domain-specific abstractive summarization models and general-domain LLMs as well as extractive summarization models over two sets of legal case judgements -- from the United Kingdom (UK) Supreme Court and the Indian (IN) Supreme Court -- and evaluated the quality of the generated summaries.  
We also perform experiments on a third dataset of legal documents of a different type -- Government reports from the United States (US). 
Results show that abstractive summarization models and LLMs generally perform better than the extractive methods as per traditional metrics for evaluating summary quality. However, detailed investigation shows the presence of inconsistencies and hallucinations in the outputs of the generative models, and we explore ways to reduce the hallucinations and inconsistencies in the summaries. 
Overall, the investigation suggests that further improvements are needed to enhance the reliability of abstractive models and LLMs for legal case judgement summarization. At present, a human-in-the-loop technique is more suitable for performing manual checks to identify inconsistencies in the generated summaries.}




\maketitle





\maketitle

\section{Introduction}

Summarizing legal case judgements is a practical and challenging task, given the complicated and lengthy nature of the legal judgements.
Typically, legal judgements are manually summarized by legal practitioners; in fact, many legal information systems offer case summaries or headnotes written by legal practitioners.
To reduce the pressure on humans, there have been several attempts towards automatic summarization of legal judgements.
Traditionally, extractive summarization methods -- which extract important sentences/parts from the input document -- have been used to summarize legal case judgements~\cite{deroy2023ensemble,bhattacharya2021incorporating,polsley-etal-2016-casesummarizer,10.1145/3322640.3326728,deroy2024artificialintelligenceailegal}. 
But recently there has been increasing interest in using \textit{abstractive} summarization models  because these models are considered to generate more `natural' and `coherent' summaries.
Hence there are some recent works that train abstractive summarization models on (legal document, summary) pairs~\cite{shukla2022legal,feijo2023improving}.
Legal domain-specific versions of pre-trained abstractive summarization models (e.g., Longformer~\cite{beltagy2020longformer}, Pegasus~\cite{zhang2020pegasus}) have also been released, such as Legal-LED\footnote{\url{https://huggingface.co/nsi319/legal-led-base-16384}} and Legal-Pegasus\footnote{\url{https://huggingface.co/nsi319/legal-pegasus}}. 

Additionally, the recent times have seen the advent of general-domain Large language models (LLMs) which can be used for a wide variety of NLP tasks including summarization, translation, question answering, and so on~\cite{teubner2023welcome}. 
These LLMs have the advantage that they can be applied for summarization of any document, including legal documents, without further training. In fact, LLMs have been used recently for news document summarization~\cite{zhang2023benchmarking}.
So the question arises how well the pre-trained abstractive summarization models and LLMs (like ChatGPT and Davinci) can perform in the task of legal case judgement summarization. We attempt to answer this question in this paper.

In this work, we apply various legal domain-specific abstractive summarization models (such as Legal-LED and Legal-Pegasus) as well as general-domain LLMs on two datasets of legal case judgements and their summaries -- 
(1)~the UK-Abs dataset (a dataset of United Kingdom case judgements), and (2)~IN-Abs dataset (a dataset of Indian case judgements).
For comparison, we also apply extractive summarization models on the same datasets.
We also perform experiments on a third dataset from the legal domain -- (3)~the GOVREPORT dataset (consisting of long reports published by the U.S. Government Accountability Office). 
We compute a large number of summary quality
metrics for all the models, 
including traditional metrics
such as ROUGE~\cite{lin2004rouge}, METEOR~\cite{banerjee2005meteor} and BERTScore~\cite{zhang2019bertscore} (that match model-generated summaries with gold standard summaries) as well as
metrics for quantifying the consistency of summaries
with respect to the original document such as SummaC~\cite{laban2022summac}, NEPrec~\cite{deroy2023ready}, and Numprec~\cite{deroy2023ready}. 

Our results show that abstractive summarization models and LLMs usually perform better than the extractive summarization methods in terms of the conventional summary quality metrics (see Section \ref{sec:results}). 
However, summaries generated by the abstractive models and LLMs often contain inconsistencies and hallucinations~\cite{10.1145/3571730}. 
We manually analyze the summaries generated by these models to report examples of such hallucinations and inconsistencies in the summaries (see Section~\ref{sec:inconsistency_examples}). 
We also explore ways of reducing such inconsistencies in the generated summaries, by domain-specific fine-tuning of abstractive summarization models, suitable prompting of LLMs and a semantic similarity-based approach. 
We also provide a human evaluation (by Law students) of summaries generated by some of the models. 
Our contributions in this work are as follows:\\
(1)~We experiment with general-domain LLMs (and several different prompts), legal domain-specific abstractive summarization models, and extractive summarization methods for legal case judgement summarization. To our knowledge, this is the first systematic comparison of all these families of summarization models for this practically important problem.\\
(2)~We show several examples of hallucinations and inconsistencies in the summaries generated by the abstractive summarization models and LLMs (see Section~\ref{sec:inconsistency_examples}).\\
(3)~We explore several different ways to reduce hallucinations and inconsistencies in legal case summaries (see Section~\ref{sec:reduce_inconsistency}). 
In particular, we develop a simple semantic similarity-based approach that effectively reduces hallucinations while also improving the quality of the generated summaries.\\
(4)~We perform a human evaluation of the summaries generated by the best performing summarization models based on several human evaluation metrics (see Section~\ref{sec:human-eval}).\\

\section{Related Work}
\label{sec:related-works}


\noindent \textbf{Legal case Judgement and Long Document Summarization:}
Several extractive and abstractive summarization methods have been used for summarizing legal case judgments~\cite{inbook11,shukla2022legal,feijo2023improving,nigam2023nonet,nigam2023fact}. 
In particular, recent works have applied abstractive summarization models like Legal-Pegasus, Legal-LED, and BART to Indian and UK court judgements~\cite{shukla2022legal}.
To this end, supervised extractive and abstractive models have been trained / fine-tuned on (legal judgement, summary) pairs. 

Unsupervised extractive models have also been developed for this purpose, such as CaseSummarizer\cite{polsley-etal-2016-casesummarizer}, DelSumm~\cite{bhattacharya2021incorporating}, and so on.
Graphical models have also been used for the purpose of legal case judgement summarization~\cite{saravanan2006improving}. 

\vspace{2mm}
\noindent \textbf{LLMs for summarization and other tasks in different domains:}
With the recent popularity of LLMs, they have been applied for text summarization as well. 
A work on news document summarization~\cite{zhang2023benchmarking} tried to investigate the performance of multiple LLMs based on the type of prompts and LLM size. The study found that prompting techniques play a critical role in determining the quality of the summaries. 
A recent work~\cite{10.1145/3551349.3559555} on code summarization using GPT used zero-shot and few-shot capabilities of LLMs for project-specific code summarization. 
A work on book summarization~\cite{chang2023booookscore} tried to develop novel methods to summarize long books using foundational models like GPT-4 and Turbo-GPT-3.5. They divide the entire book into fragments and then follow a hierarchical strategy of divide and conquer to summarize the entire corpus of a long book.
LLMs have also been used for comment summarization of videos and topic matching for videos~\cite{dimarco2023llm}.
Summarization of long meetings has also been attempted using LLMs like GPT-4 and GPT-3.5 and encoder-decoder models like BART~\cite{schneider2023team}.
There is a work~\cite{maity2024novel} on distractor generation in MCQs where a multi-stage prompting approach delivers better results than a single-stage prompting approach. There are several works on question generation~\cite{maity2023harnessing,maity2024exploring,maity2024effective} for school level subjects using LLMs. There are several works on classification~\cite{deroy2021multi}, summarization~\cite{deroy2023prompted}, low resource language learning~\cite{maity2024ready}, bias detection in texts~\cite{deroy2023questioning}, etc which uses LLMs

However, to our knowledge, LLMs have not been much tried for legal document summarization. We attempt to fill this gap in the present paper.  

\vspace{2mm}
\noindent \textbf{Hallucinations, Inconsistencies and Unfaithfulness in Large language models and Extractive Summarization models:}
Hallucinations are a serious pitfall of LLMs and other natural language  generation models (including abstractive text summarization models) as they may generate content which is non-factual and sometimes irrelevant to the context~\cite{hallucination-survey}. Particularly, in domains such as medical/healthcare and legal, hallucinations can lead to serious outcomes. 
For instance, a recent work~\cite{ahmad2023creating} focused on the use of LLMs in the medical domain where there is a reluctance of professionals to use LLMs due to hallucinations and generation of inconsistent content.
So this work focused on strategies and techniques to mitigate hallucinations in LLMs so that these models can be suitably used by medical professionals without any issues. 
Note that unfaithfulness has also been observed in \textit{extractive} summaries~\cite{zhang-etal-2023-extractive}. This study categorized five types of broad unfaithfulness issues that can arise in extractive summaries, extending beyond mere non-entailment. These issues include incorrect co-reference, incomplete co-reference, incorrect discourse, incomplete discourse, and other forms of misleading information.

In the present work, we point out several examples of hallucinations by LLMs and other abstractive summarization models while summarizing case judgements, and then explore ways to reduce such hallucinations.

\vspace{2mm}
\noindent 

\noindent \textbf{Segmentation for long document summarization:} 
Legal documents are mostly long, while most LLMs have a limit on the size of the input/prompt that can be given at once. 
So, in this work, we divide long legal documents into chunks where every chunk is semantically related, and then summarizing one chunk at a time. 
Such chunking or segmentation-based methods for summarization of long documents have been used in prior works. 
For instance, semantic segmentation  was used in~\cite{moro2022semantic} to summarize long legal documents in low-resource settings. They partitioned a long input into semantically coherent chunks, allowing transformers to summarize very long documents without truncation by summarizing each chunk and then concatenating the results.
Again, a multi-stage summarization framework for long 	
dialogues and documents was developed in~\cite{zhang-etal-2022-summn} where the source data was broken into chunks so that it does not exceed the token limit of the summarization models. 
Moro et al.~\cite{MORO2023126356} developed a novel align-then-abstract representation learning model for low resource summarization, where they jointly trained a summarizer and a segmenter by maximizing the alignment between the chunk-segment pairs in the output from text segmentation. 
In another work, Moro et al.~\cite{moro2023efficient} developed another model for long document summarization, where the model processes lengthy inputs by dividing them into multiple text fragments, allowing it to store and compare the current chunk with previous ones. This approach lets the model understand the full context of a document while using a fixed amount of GPU memory.

\vspace{2mm}
\noindent \textbf{Present work as an extension of our prior work:} The present work is a much extended version of our prior work~\cite{deroy2023ready}. 
The present work extends the work in~\cite{deroy2023ready} in the following ways:
(1)~The work~\cite{deroy2023ready} considered only one dataset of Indian legal documents, while the present work adds extensive experiments on two new datasets of UK legal documents and the GOVREPORT dataset.
(2)~The present work explores many variations of the LLMs --ChatGPT, Davinci, Llama2-70b and GPT-4-Turbo -- which were not explored in~\cite{deroy2023ready}. 
Also some new extractive summarization methods like PACSUM~\cite{zheng-lapata-2019-sentence} and HipoRank~\cite{dong2021discourse} are applied in this work. 
(3)~The present work additionally explores different ways of reducing hallucination/inconsistency in the summaries generated by abstractive summarization models and LLMs. 
We also present a novel semantic similarity-based approach for reducing hallucinations.
(4)~The present work includes a human evaluation of the summaries generated by different methods.

\section{Datasets for Legal Judgement Summarization}
\label{sec:datasets}

For most of the experiments in this paper, we use two datasets of court case judgements and their gold standard summaries, that are described below.  

\vspace{2mm}
\noindent {\bf The IN-Abs dataset}, obtained from the prior work~\cite{shukla2022legal}, consists of Indian Supreme Court Case Judgements collected from the website of the Legal Information Institute of India.\footnote{\url{http://www.liiofindia.org/in/cases/cen/INSC/}}
It consists of 7,130 (legal judgement, summary) pairs. The gold standard / reference summaries present in this dataset are the ``headnotes'' written by legal experts recruited by the Legal Information Institute Of India.\footnote{We use the terms `reference summaries' and `gold standard summaries' interchangeably in this paper.} 
As an example, one of the documents used in this dataset can be seen at \url{https://indiankanoon.org/doc/1801104/}. The page consists of some meta-data (e.g., names of the petitioner, the respondent, the judges, etc.), the ``HEADNOTE", and the ``JUDGMENT''. The dataset was created by collecting such pages, and extracting the headnote part and the judgement part. The headnote part serves as the gold standard summary of the judgement.

The dataset is split into a training set consisting of 7030 (legal judgement, summary) pairs and the test set consisting of 100 (legal judgement, summary) pairs.
Table~\ref{tab:IN-Abs-stats} shows the dataset statistics for the IN-Abs dataset.

Following~\cite{grusky-etal-2018-newsroom,shen2022multi}, the coverage and density of the legal judgements with respect to the gold-standard summaries are 0.35 and 1.67 respectively for this dataset.

\vspace{2mm}
\noindent {\bf The UK-Abs dataset}, which is also  obtained from~\cite{shukla2022legal}, consists of a collection of 793 legal case documents from the United Kingdom's Supreme Court website\footnote{\url{https://www.supremecourt.uk/decided-cases/}}. Along with the judgements, the website also provides official press summaries for legal cases, which are considered as the reference (gold standard) summaries. 
As an example, one of the cases in this dataset is available at \url{https://www.supremecourt.uk/cases/docs/uksc-2015-0063-judgment.pdf} and its press summary is available at \url{https://www.supremecourt.uk/cases/docs/uksc-2015-0063-press-summary.pdf}. The dataset was created by extracting the text from such PDF documents. 

Out of the 793 (legal document, summary) pairs, 100 pairs were used as the test set and 693 pairs were used as the training dataset.
Table~\ref{tab:UK-Abs-stats} shows the dataset statistics for the UK-Abs dataset.
Following~\cite{grusky-etal-2018-newsroom,shen2022multi}, the coverage and density of the legal judgements with respect to the gold-standard summaries in this dataset are 0.23 and 1.53 respectively.

\begin{table}[tb]
    \centering
\begin{tabular}{|p{0.1\columnwidth}|p{0.15\columnwidth}|p{0.25\columnwidth}|p{0.3\columnwidth}|}
\hline
 & \textbf{Nos. of documents} & \textbf{Average nos. of words per document} & \textbf{Average nos. of words per gold-standard summary} 
\\ \hline 
Train set & 7030 & 4368.49 &  839.75 \\ \hline
Test set & 100 & 4782.71 & 932.01 \\ \hline

\end{tabular}
\caption{\textbf{Dataset statistics for IN-Abs dataset.} To get the average number of words per document (or summary), we first calculate the number of words in every document (summary). Then we add up the number of words from all documents (summaries), and divide the total word-count by the number of documents (summaries).}
\label{tab:IN-Abs-stats}
\end{table}

\begin{table}[tb]
    \centering
\begin{tabular}{|p{0.1\columnwidth}|p{0.15\columnwidth}|p{0.25\columnwidth}|p{0.3\columnwidth}|}
\hline
 & \textbf{Nos. of documents} & \textbf{Average nos. of words per document} & \textbf{Average nos. of words per gold-standard summary} 
\\ \hline 
Train set & 693 & 12812.01 &  1097.38 \\ \hline
Test set & 100 & 14476.49 & 1095.49 \\ \hline

\end{tabular}
\caption{\textbf{Dataset statistics for UK-Abs dataset}. 
The average number of words per document or summary is computed as stated in Table~\ref{tab:IN-Abs-stats}.}

\label{tab:UK-Abs-stats}
\end{table}

\section{Summarization Models}

In this study, we explore a variety of summarization models, categorizing them into three main categories -- (1)~extractive summarization models, 
(2)~general purpose LLMs applied as summarizers, and
(3)~legal domain-specific abstractive summarization models.
This section describes all the models in detail.

\subsection{Extractive summarization models}

We try 5 different extractive summarization methods.
Three of these methods -- CaseSummarizer~\cite{polsley-etal-2016-casesummarizer}, BertSum~\cite{liu2019fine}, SummaRunner~\cite{nallapati2017summarunner} -- were observed to perform well for legal case judgement summarization in the prior work~\cite{deroy2021analytical}.
Additionally, we apply two recent methods PACSUM~\cite{zheng-lapata-2019-sentence} and HipoRank~\cite{dong2021discourse}.

\vspace{1mm}
\noindent \textbf{CaseSummarizer}~\cite{polsley-etal-2016-casesummarizer} -- This unsupervised method ranks sentences based on a TF-IDF matrix which is created using the corpus of legal judgements. CaseSummarizer adjusts sentence scores based on dates, entities, and closeness to section headings. The implementation of this model has been taken from \url{https://github.com/Law-AI/summarization/tree/aacl/extractive/CaseSummarizer}.

\vspace{1mm}
\noindent
\textbf{BertSum}~\cite{liu2019fine} - BertSum is a supervised summarization model which is based upon a variant of the BERT model. The sentences which are present in the gold standard summary are considered extremely important by the summarization algorithms. 
Hence BertSum is trained to perform a binary classification task of labeling sentences of the source document as summary-worthy or not. 
The implementation of this model has been used from-\url{https://github.com/nlpyang/BertSum}.

For the IN-Abs dataset, we use the training dataset of IN-Abs of 7030 (legal document, summary) pairs to train the BertSum model.
For the UK-Abs dataset, we use the training dataset of UK-Abs of 693 (legal document, summary) pairs to train the BertSum model.
Since the gold standard summaries in IN-Abs and UK-Abs are abstractive in nature, we take a sentence from a gold-standard summary and compare it with all sentences in the corresponding source judgement. Based on~\cite{moro2023graph}, we take the 3 sentences with highest Rouge-2 F1 from the main judgement to be used as a part of the reference pseudo-extractive summary for training the model. We perform this step for every sentence in the abstractive gold-standard summary and we take a union of all the corresponding sentences obtained from the main judgement to be labelled as '1' for the binary classification task. All other sentences in the main judgment are labelled as '0' to train the BertSum model.

\vspace{1mm}
\noindent 
\textbf{SummaRunner}~\cite{nallapati2017summarunner} -- SummaRunner is a supervised summarization model which is based on the Recurrent Neural Network architecture. 
The implementation of this model has been used from \url{https://github.com/hpzhao/SummaRuNNer}.
Similar to BertSum (described above), this supervised model is also trained to perform a binary classification task of labeling sentences of the source document as summary-worthy or not. 
We train the SummaRunner model in exactly the same way over the IN-Abs and UK-Abs training sets, as described above for BertSum.


\vspace{1mm}
\noindent \textbf{PACSUM~\cite{zheng-lapata-2019-sentence}} -
PACSUM (Position-Augmented Centrality based Summarization) is an extractive summarization method that enhances BERT by integrating sentence position information. It uses BERT to generate contextual embeddings, incorporates sentence positions into the scoring mechanism, constructs a similarity graph with position-adjusted weights, ranks sentences based on these weights, and then selects top-ranked sentences to form the summary. This approach results in more relevant and informative summaries by combining BERT's contextual understanding with position-aware scoring.
The github link for the code implementation is taken from-\url{https://github.com/mswellhao/PacSum}

\vspace{1mm}
\noindent \textbf{HiPoRank~\cite{dong2021discourse}} -
HiPorank (Hierarchical and
Positional Ranking model) is an extractive summarization algorithm that improves summary quality by considering hierarchical structure and positional importance. It organizes documents into sections, paragraphs, and sentences to understand context, prioritizes sentences at the beginning or end of sections/paragraphs, and represents sentences as graph nodes with edge weights adjusted for hierarchy and position. The algorithm scores and ranks sentences using this data and selects top-ranked sentences to form a coherent, comprehensive summary, resulting in more relevant and coherent summaries.
The implementation of this model has been taken from -\url{https://github.com/mirandrom/HipoRank}.

\subsection{General Purpose LLMs} \label{sub:llms-for-summarization}

We try out the following Large language models (LLMs).
All the LLMs take as input a `prompt' and generate text as a `response'. Specifically for the summarization task, the prompt consists of (i)~the text to be summarized,
which we refer to as \verb|<text to summarize>| and (ii)~an `instruction' that tells the model that the input text has to be summarized.
Note that the LLMs have a limit on the maximum number of tokens possible in a prompt+response, as stated below.

\vspace{2mm} \noindent 
\textbf{Text Davinci-003} is an advanced transformer-based LLM with 175 billion parameters. Though detailed information on the exact sources of the training data are not publicly available, 
it is known that the model has been trained on a massive text dataset using a combination of supervised and unsupervised learning methods. 
Text-Davinci-003 has a maximum prompt+response length of 4096 tokens.

\vspace{2mm} \noindent 
\textbf{Turbo-GPT-3.5} (popularly known as ChatGPT) is a cost-effective LLM, said to have approximately 154 billion parameters, which is close to Text-Davinci-003 in terms of performance. This LLM has been trained on a diverse range of text data, including
web pages, books, scientific articles, and other sources of human-written text including chats, using a combination of supervised and reinforcement learning methods.
 Turbo-GPT-3.5 has two different variations -(i)~the original version with maximum prompt+response length of 4096 tokens and (ii)~a version with a longer  prompt+response length of 16,384 tokens, popularly called `chatgpt-16k'.

\vspace{2mm} \noindent 
\textbf{GPT-4 Turbo} is one of the most powerful LLMs available today. The model can handle both text and images. It has a long context length of 128K due to which we can feed long documents into these models. The model has advanced reasoning and broader general knowledge capabilities.

\vspace{2mm} \noindent 
\textbf{Llama2-70b} is an open-source generative pretrained and fine-tuned text model with 70 billion parameters. It has a relatively long context length of 4096 tokens.

\vspace{2mm} \noindent 
We try the Davinci, GPT-3.5 and GPT-4 models using the OpenAI API.\footnote{\url{https://platform.openai.com/docs/api-reference/completions}}
The Llama2-70b model implementation is obtained from Huggingface.\footnote{\url{https://huggingface.co/meta-llama/Llama-2-70b-hf}}
All the LLMs were used in `zero-shot' mode, without any in-context learning or fine-tuning.

\subsubsection{Variations of Text-Davinci-003} 

We try different prompts with the model, leading to the following variations:

\noindent (i) \textbf{Davinci-summ:} For this variant, the prompt is ``\verb|<text to summarize>| Summarize the document in \verb|<YY>| words'' where $YY$ is a number  representing the target length of the output summary in number of words. How the value of $YY$ is decided is explained later in the section.

\noindent (ii) \textbf{Davinci-tldr:} For this variant, the prompt is ``\verb|<text to summarize> Tl;Dr|''. We first pass the document followed by the ``Tl;Dr'' identifier which is an identifier for summarization.

\noindent (iii) \textbf{Davinci-explicit:} For this variant, we use a more explicit prompt -- ``Your task is to summarize the following document in at most \verb|<YY>| words. The document to be summarized is given within \verb|<>|. Document to summarize - \verb|<text to summarize>|''.

\noindent (iv) \textbf{Davinci-hybrid:}
LLMs like Text-Davinci-003 have constraints on the length of prompt+response, which is 4096 tokens at most. 
So the idea behind this extractive-abstractive hybrid summarization technique is that an extractive summarization method is first used to filter out some key information present in the main judgement. 
In the second step, the key information is provided to the LLM to summarize the key information as an abstractive summary. 

Since CaseSummarizer performed the best amongst the extractive summarization methods (as will be seen later in the paper), we choose the CaseSummarizer method to generate the extractive summary in the first step.
First, we create an extractive summary of 1500 words (approx 2000 tokens) or less using the CaseSummarizer method.
The sentences chosen for inclusion in the extractive summary are placed in the order in which they are present in the source document. Next we create an abstractive summary from this extractive summary using the prompt ``\verb|<text to summarize>| Summarize the document in \verb|<YY>| words'' where YY is a number that represents the target length of the output summary in number of words.\\

\subsubsection{Variations of Turbo-Gpt-3.5 (ChatGPT)} 

Similar to what we tried for Davinci, we try  different variations of the ChatGPT model:

\noindent (i) \textbf{Chatgpt-summ:} For this variant, the prompt is ``\verb|<text to summarize>| Summarize the document in \verb|<YY>| words'' where $YY$ is a number representing the target length of the output summary in number of words. How the value $YY$ is decided is explained later in the section.

\noindent (ii) \textbf{Chatgpt-tldr:} For this variant, the prompt is ``\verb|<text to summarize> Tl;Dr|''. We first pass the ``Tl;Dr'' identifier followed by the text to be summarized.

\noindent (iii) \textbf{Chatgpt-explicit:} For this variant, the prompt is ``Your task is to summarize the following document in at most \verb|<YY>| words. The document to be summarized is given within \verb|<>|. Document to summarize - \verb|<text to summarize>|''.

\noindent (iv) \textbf{Chatgpt-hybrid:}
The original ChatGPT model also has a constraint on the maximum prompt+response length which is 4096 tokens. 
So the idea behind this extractive-abstractive hybrid summarization technique is to first use an extractive summarization method to filter out the key information present in the main judgement, and then to generate an abstractive summary of the key information using the LLM. As before, we use CaseSummarizer to generate the extractive summary of at most 1500 words. All sentences in the extractive summary are placed in the order in which they appear in the document.
Then we create an abstractive (target) summary from this extractive summary using the prompt ``\verb|<text to summarize>| Summarize the document in \verb|<YY>| words'' where YY is a number that represents the target length of the output summary in number of words.

\noindent (v) \textbf{chatgpt-16k-long:}
Here we use the chatgpt-16k variant which has an input+response length of 16k tokens. This variant can be fed with longer inputs, hence it is natural to try out this variant for summarization of long legal documents. 
For this variant, we use the prompt ``Summarize the document in \verb|<YY>| words: \verb|<text to summarize>|'' (the prompt is the same as in Chatgpt-summ) where $YY$ is a number representing the target length of the output summary in number of words.


\subsubsection{Variations of GPT-4-Turbo (gpt4)} 


The various GPT4 variations we experiment with are:-

\noindent (i) \textbf{gpt4-summ:} For this variant, the prompt is ``\verb|<text to summarize>| Summarize the document in \verb|<YY>| words'' where $YY$ is a number representing the target length of the output summary in number of words. How the value $YY$ is decided is explained later in the section.

\noindent (ii) \textbf{gpt4-tldr:} For this variant, the prompt is ``\verb|<text to summarize>| Tl;Dr''. We first pass the ``Tl;Dr'' identifier followed by the text to be summarized.

\noindent (iii) \textbf{gpt4-explicit:} For this variant, the prompt is ``Your task is to summarize the following document in at most \verb|<YY>| words. The document to be summarized is given within $<>$. Document to summarize - \verb|<text to summarize>|''.

\noindent (iv) \textbf{gpt4-hybrid:}
In this extractive-abstractive hybrid summarization technique, we first use an extractive summarization method to filter out the key information present in the main judgement, and then to generate an abstractive summary of the key information using the LLM. As before, we use CaseSummarizer to generate the extractive summary of at most 1500 words. All sentences in the extractive summary are placed in the order in which they appear in the document.
Then we create an abstractive (target) summary from this extractive summary using the prompt ``\verb|<text to summarize>| Summarize the document in \verb|<YY>| words'' where $YY$ is a number that represents the target length of the output summary in number of words.

\subsubsection{Variations of Llama2-70b} 

Similar to the variations for the other LLMs, we try the following variations of Llama2-70b :-

\noindent
(i) \textbf{llama-summ:} For this variant, the prompt is ``\verb|<text to summarize>| Summarize the document in \verb|<YY>| words'' where $YY$ is a number representing the target length of the output summary in number of words. How the value $YY$ is decided is explained later in the section.

\noindent
(ii) \textbf{llama-tldr:} For this variant, the prompt is ``\verb|<text to summarize>| Tl;Dr''. We first pass the ``Tl;Dr'' identifier followed by the text to be summarized.

\noindent
(iii) \textbf{llama-explicit:} 
For this variant, the prompt is ``Your task is to summarize the following document in at most \verb|<YY>| words. The document to be summarized is given within \verb|<>|. Document to summarize - \verb|<text to summarize>|''.

\noindent
(iv) \textbf{llama-hybrid:}
In this extractive-abstractive hybrid summarization technique we first use an extractive summarization method (CaseSummarizer) to filter out the key information present in the main judgement, and then to generate an abstractive summary of the key information using the LLM. As before, we use CaseSummarizer to generate the extractive summary of at most 1500 words. All sentences in the extractive summary are placed in the order in which they appear in the document.
Then we create an abstractive (target) summary from this extractive summary using the prompt ``\verb|<text to summarize>| Summarize the document in \verb|<YY>| words'' where $YY$ is a number that represents the target length of the output summary in number of words.

\subsection{Legal domain-specific abstractive summarization models}

We tried two different legal domain-specific abstractive summarization models namely Legal-Pegasus (abbreviated as LegPegasus) and Legal-LED (abbreviated as LegLED). 

\vspace{2mm}
\noindent {\bf LegPegasus:}
Pegasus (formally, \texttt{google/pegasus-cnn\_dailymail}) is a general-purpose abstractive summarization model developed by Google.
This model was fine-tuned on 
the `sec-litigation-releases' dataset -- consisting of 2,700 litigation releases and complaints related to civil lawsuits in various courts in the United States of America (USA) along with their summaries --  to develop the LegPegasus model designed specifically for abstractive summarization in the legal domain. LegPegasus can be accessed at \url{https://huggingface.co/nsi319/legal-pegasus} and has a maximum input token length of 1024 tokens. 

\vspace{2mm}
\noindent {\bf LegLED:}
This model is based on the Longformer architecture, a transformer-based neural network architecture designed to process long sequences of text efficiently. The LegLED model has also been fine-tuned on the same `sec-litigation-releases' dataset consisting of 2700 (legal document, summary) pairs related to civil lawsuits in various courts in the USA. 
It is also specifically designed for summarization in the legal domain. The LegLED model can be accessed at \url{https://huggingface.co/nsi319/legal-led-base-16384}. The model  has a maximum input token length of 16,384.

\subsection{Chunking of long legal documents}

As stated earlier, Text-Davinci-003 and Turbo-Gpt-3.5 have a token limit of maximum 4,096 for (prompt+generated text). 
One token is approximately $\frac{3}{4}$ words, i.e., 1000 tokens correspond to around 750 words.\footnote{Detailed explanation about tokens available at \url{https://help.openai.com/en/articles/4936856-what-are-tokens-and-how-to-count-them}}
Also, legal domain-specific abstractive summarization models like LegPegasus have a maximum input token length of 1,024.

Recall from Section~\ref{sec:datasets} that the average length of UK case judgements in the UK-Abs test dataset is 14,476 words and that of the Indian case judgements in the IN-Abs test dataset is 4,782 words. 
Thus, these legal case judgements are often much longer than what can be input into the summarization models/LLMs at once, and hence we have to follow a divide-and-conquer approach with the long legal documents. 
Our strategy involves chunking long legal documents into smaller segments or chunks of at most $K$ words (where $K$ can be 1,024 or 2,048 or higher), and each chunk is passed individually into the summarization models to obtain the output summary. 
Then the summaries generated for all the chunks (of a given document) are appended together in the order in which the chunks appear in the main document, to form the final output summary for the legal judgement. 
For legal documents with a length of lesser than 1024 words, the summary is obtained at once by passing the entire document through the summarization models.

For the ChatGPT and Davinci models, we experiment with $K$ = 1,024 and 2,048. 
For the chatgpt-16k model, we divide a legal judgement into longer chunks of length $K$ = 8192 words (10,922 tokens approx). 
Note that almost all documents in the IN-Abs test set and a large majority of documents in the UK-Abs test set are actually shorter than 8,192 words; hence the summary is obtained at once by passing the entire document through the chatgpt-16k model.

\vspace{2mm}
\noindent \textbf{Deciding the target summary length of a chunk:}
When text is input to an LLM such as ChatGPT, we also need to specify a hyperparameter namely `max tokens' to specify the maximum desired length of the summary to be generated (in terms of number of words).

Let $\lvert Doc \rvert$ be the length of an input document, and let $\lvert Exp \rvert$ be the length of its reference summary written by human experts. 
Assume that the document is split into a series of chunks, each of size $K$ words.
Then the target summary length to be generated by the model for each model is $K \times \lvert Exp \rvert / \lvert Doc \rvert $ words. 
In other words, we use the LLMs / summarization models 
to summarize every chunk considering the same compression ratio as for the whole document and the reference summary.

There is an inherent limitation in this simple method. In reality, all parts of the document are not equally important, hence different chunks should ideally be allocated different lengths in the summary. In contrast, this method allocates the same length in the summary for all chunks.
However, there is no simple way of knowing the relative importance of different chunks in a legal case judgement; hence we follow the simple method of assuming the same compression ratio for every chunk.

\section{Performance Metrics}
\label{sec:metrics}

We compare the
quality of summaries generated by the different methods
along two aspects -- 
(1)~their match with the gold standard or reference summaries, and 
(2)~their consistency with the input documents.

\subsection{Match with gold-standard summaries}

Here we measure the match of an algorithm-generated summary for a document with the gold standard / reference summary of the same document. The following metrics are used to measure the performance of the summarization models. 

\vspace{1mm}
\noindent \textbf{ROUGE:}
The term `ROUGE' stands for ``Recall-Oriented Understudy for Gisting Evaluation''~\cite{lin2004rouge}. 
ROUGE is a family of metrics used for the automatic evaluation of machine-generated text, particularly in the context of  text summarization. Here we measure the Rouge-score between the expert-written summaries and model-generated summaries. Specifically, we calculate Rouge-2 precision, recall and F1 scores, which evaluate the \textit{bigram match}, and Rouge-L precision, recall, F1 scores which measure the Longest Common
Subsequence-based match between the model-generated summaries
and the gold standard summaries.


\vspace{1mm}
\noindent \textbf{METEOR}~\cite{banerjee2005meteor} is a metric used for the automatic evaluation of machine translation and summarization output. 
It was designed to address some limitations of other metrics like BLEU by incorporating explicit word order information and considering synonyms and stemming.
This metric measures the unigram overlap between expert-written summaries and model-generated summaries.

\vspace{1mm}
\noindent \textbf{BERTSCORE}~\cite{zhang2019bertscore} is a popular metric for evaluating the quality of machine-generated text, especially in the context of natural language processing (NLP) tasks such as text summarization, machine translation, and question answering. Unlike traditional evaluation metrics like ROUGE, which rely on exact matches or n-gram overlaps, BERTScore takes into account the \textit{semantic similarity} between the reference (ground truth) summary and the model generated summary.

\vspace{1mm}
\noindent 
To implement these metrics, we employ the SummEval package  available at \url{https://github.com/Yale-LILY/SummEval}, which is a well-known toolkit used for the evaluation of summarization.

\subsection{Metrics for consistency of summaries}
\label{consistency}

Now we discuss three metrics used to assess the consistency of model-generated summaries.
These metrics compare a model-generated summary with the original document and estimate how consistent the summary is with the document. 
All these metrics give a score in the range [0, 1]; the higher the
score, the more consistent is the summary.

\vspace{2mm}
\noindent 
\textbf{SummaC}~\cite{laban2022summac} -- This metric utilizes Natural Language Inferencing (NLI) to determine the logical relationship between sentences. 
The NLI task involves determining the relationship between two sentences. One of the sentences
is considered as a `hypothesis' and the other sentence is
considered as a `premise'. 
Typically, a NLI model will give a score representing how likely the hypothesis sentence is to logically follow from the premise sentence.

Given a (document, summary) pair, the SummaC metric~\cite{laban2022summac}
computes NLI scores for each sentence in the model-generated summary, indicating the likelihood that the sentence logically follows from the sentences in the original document. 
Lower NLI scores for
a particular sentence $s$ in the summary suggest a higher mismatch between this sentence and the document, and hence the potential presence of hallucinated information. 
The NLI scores of individual sentences in the summary are combined to give a single SummaC score for the given (document, summary) pair. 
A higher SummaC score indicates greater consistency between the model-generated summary and the original document.
We use the standard implementation of SummaC available at \url{https://github.com/tingofurro/summac}.

\vspace{2mm}
\noindent 
\textbf{NumPrec} - Numbers play a significant role in legal case judgements, such as dates, statute identifiers, monetary values, etc. The NumPrec metric measures the fraction of numbers present in the model-generated summary that also appear in the source document. 
We rely on the standard Python library for number identification.

\vspace{2mm}
\noindent 
\textbf{NEPrec} - Named Entities are important for legal case judgements, and changes in entities like a person's name or organization's name can result in information loss and potential misrepresentation. 
The NEPrec metric measures the fraction of named entities present in the model-generated summary that also exists in the original document. 
We use the Spacy toolkit\footnote{\url{https://spacy.io/}} to detect named entities in original documents as well as the summaries.
It is worth noting that the accuracy of the NEPrec metric depends on the accuracy of the toolkit used to identify named entities.

\section{Results} \label{sec:results}

In this section, we study the performance of various summarization models on the two datasets.

\subsection{Selecting the chunk size for different LLMs}

\begin{table*}[tb]
\centering
\small
\scalebox{0.9}
{
\begin{tabular}{l|lll|lll|ll}
\hline
\textbf{Model} & \textbf{R2-P} & \textbf{R2-R} & \textbf{R2-F1} & \textbf{RL-P} & \textbf{RL-R} & \textbf{RL-F1}  & \textbf{ME} & \textbf{BS}  
\\ \hline

\multicolumn{9}{|c|}{\textbf{Chatgpt}}  \\ \hline

Chatgpt-summ (1024)  & 0.196 & \textbf{\textcolor{blue}{0.173}} & \textbf{\textcolor{blue}{0.181}} & 0.236 & \textbf{\textcolor{blue}{0.208}} & \textbf{\textcolor{blue}{0.218}} & \textbf{\textcolor{blue}{0.196}} & \textbf{\textcolor{blue}{0.627}} \\ \hline

Chatgpt-summ (2048)  & 0.187 & 0.151 & 0.163 & 0.208 & 0.191 & 0.200  & 0.185  & 0.617 \\ \hline

Chatgpt-16k-long (8192) & \textbf{\textcolor{blue}{0.231}} & 0.127 & 0.149 & \textbf{\textcolor{blue}{0.319}} & 0.176 & 0.206 & 0.147 & 0.615  \\ \hline

\multicolumn{9}{|c|}{\textbf{Davinci}}  \\ \hline

Davinci-summ (1024)  & \textbf{\textcolor{blue}{0.220}} & \textbf{\textcolor{blue}{0.179}} & \textbf{\textcolor{blue}{0.195}} & \textbf{\textcolor{blue}{0.251}} & \textbf{\textcolor{blue}{0.205}} & \textbf{\textcolor{blue}{0.223}} & 0.191 & 0.624 \\ \hline 

Davinci-summ (2048)  & 0.190 & 0.167 & 0.174 & 0.223 & 0.187 & 0.192 & \textbf{\textcolor{blue}{0.218}}  & \textbf{\textcolor{blue}{0.627}}   \\ \hline

\multicolumn{9}{|c|}{\textbf{Llama}}  \\ \hline

Llama-summ (1024)   & \textbf{\textcolor{blue}{0.186}} & \textbf{\textcolor{blue}{0.144}} & \textbf{\textcolor{blue}{0.163}} & \textbf{\textcolor{blue}{0.219}} & \textbf{\textcolor{blue}{0.187}} & \textbf{\textcolor{blue}{0.192}} & 0.152 & \textbf{\textcolor{blue}{0.621}}  \\ \hline 

Llama-summ (2048)  & 0.180 & 0.140 & 0.158 & 0.213 & 0.180 & 0.182 & \textbf{\textcolor{blue}{0.166}}  & 0.600   \\ \hline




\end{tabular}
}
\caption{Summarization performances of ChatGPT, Davinci, and Llama for different chunk sizes on the IN-Abs dataset. Chunk sizes are stated within parentheses. The same prompt is used for all models. Best value of each metric for each family of models is in blue-bold font.}
\label{tab:comparative-metrics-rouge}
\end{table*}

First we compare the performances of Chatgpt, Davinci, and Llama with different chunk sizes, to check which chunk size gives the best performance. We perform these experiments on the IN-Abs dataset.

Table~\ref{tab:comparative-metrics-rouge} shows the ROUGE, METEOR, BERTScore metrics for chatgpt-summ, chatgpt-16k-long, davinci-summ, and llama-summ, using different chunk sizes. The same prompt is used for all the different variations of Chatgpt, Davinci, and Llama, as stated in Section~\ref{sub:llms-for-summarization}. 
The highest value for every metric is shown in blue-bold.

For all the three LLMs (chatgpt, davinci and Llama), 
the best results for most metrics are obtained with chunk-size 1024. 
This is possibly because the LLMs find it difficult to capture the context when the chunk sizes become too large, thus leading to better summaries with chunk size 1024. 
However, we preferred not to use even shorter chunks, since there is a possibility of redundancy and loss of continuity/coherence if the document is broken into too many chunks. 
Hence, we perform all subsequent experiments with chatgpt, davinci, and Llama considering a chunk size of 1024 words. 

For chatgpt-16k, we will continue to use longer chunks of 8192 words.
For GPT-4 Turbo, we feed the entire document into the model since GPT-4 Turbo has a very long context length of 128K.


\subsection{Summarization results on UK-Abs dataset}

We first compare the performances of LLM-based summarization using different prompts (that were stated in Section~\ref{sub:llms-for-summarization}), and select the best LLM variants for summarization of this dataset. 
Then we compare the best LLM variants with other extractive and abstractive summarization models. 

\vspace{2mm}
\noindent {\bf Comparing LLM-based summarization variants:}
Table~\ref{tab:general-metrics-ukabs-rouge} shows the ROUGE, METEOR, and BERTScore metric for the general-purpose LLMs on the UK-Abs dataset.
Among the Chatgpt variants (all of which are run considering a chunk length of 1024 words, as stated earlier), the Rouge-2 F1 score for chatgpt-summ is higher than the other chatgpt variants.
The chatgpt-explicit variation achieves the highest score amongst all chatgpt variations for most metrics, including Rouge-2 recall, Rouge-L Recall, Rouge-L F1, METEOR and BERTScore.
Among the Davinci variants, Davinci-summ achieves the highest scores for most of the metrics. Among the gpt4 variants, gpt4-explicit achieves the highest scores for most of the metrics. Among the llama variants, llama-explicit achieves the highest scores for most of the metrics.


We observe that the `hybrid' variants have higher Rouge-2 precision and Rouge-L precision across all the model variants. This suggests that the summaries generated by the `hybrid' variants contains many of the same sequences of words (in the same order) as the reference / gold standard summaries.
Whereas, the `explicit' and `summ' variants have higher Rouge-2 recall and Rouge-L recall across all the model variants, 
suggesting that the summaries include a significant portion of the content of the reference summaries. 

An interesting observation is that the summaries created by GPT-4 Turbo are generally shorter in length, compared to the summaries created by Turbo-GPT-3.5 (Chatgpt). This is why the summaries created by Chatgpt have higher matches with the gold standard summaries, thereby having higher scores for the Recall and FI metrics. However, the summaries generated by GPT-4 achieve higher Rouge-Precision scores in most cases. 

We now compare some of the best performing LLM variants with the other types of summarization models.

\begin{table*}[tb]
\centering
\small
\scalebox{0.9}
{
\begin{tabular}{l|lll|lll|ll}
\hline

\textbf{Model} & \textbf{R2-P} & \textbf{R2-R} & \textbf{R2-F1} & \textbf{RL-P} & \textbf{RL-R} & \textbf{RL-F1} & \textbf{ME} & \textbf{BS}\\ \hline  

\multicolumn{9}{|c|}{\textbf{Llama2-70b(chunk size 1024)}} 
\\ \hline

llama-tldr  & 0.156 & 0.121 & 0.136 & 0.186 & 0.159 & 0.168 & 0.184 & 0.605 \\ \hline

llama-summ  & 0.178  & 0.126 & 0.148 & 0.234 & \textcolor{blue}{0.174} & \textcolor{blue}{0.186} & 0.196 & 0.625 \\ \hline

llama-explicit & 0.165 & \textcolor{blue}{0.129} & \textcolor{blue}{0.148} & 0.206 & 0.173 & 0.181 & \textcolor{blue}{0.200} & \textcolor{blue}{0.629}\\ \hline


llama-hybrid & \textcolor{blue}{0.210} & 0.062 & 0.104 & \textcolor{blue}{0.320} & 0.104 & 0.137 & 0.121 & 0.619 \\ \hline

\multicolumn{9}{|c|}{\textbf{GPT-4 Turbo}} 
\\ \hline

gpt4-tldr  & 0.172 & 0.134 & 0.149 & 0.203 & 0.179 & 0.187 & 0.208 & 0.629 \\ \hline

gpt4-summ  & 0.193  & 0.146 & 0.159 & 0.235 & \textcolor{blue}{0.182} & \textcolor{blue}{0.198} & 0.207 & 0.631 \\ \hline

gpt4-explicit & 0.184 & \textcolor{blue}{0.148} & \textcolor{blue}{0.168} & 0.228 & 0.179 & 0.195 & \textcolor{blue}{0.209} & \textcolor{blue}{0.643}\\ \hline


gpt4-hybrid & \textcolor{blue}{0.220} & 0.064 & 0.107 & \textbf{\textcolor{blue}{0.327}} & 0.108 & 0.148 & 0.125 & 0.626 \\ \hline

\multicolumn{9}{|c|}{\textbf{Chatgpt (chunk size 1024)}} 
\\ \hline

chatgpt-tldr  & 0.152 & 0.151 & 0.151 & 0.198 & 0.196 & 0.196 & 0.216 & 0.634 \\ \hline

chatgpt-summ  & 0.188  & 0.160 & \textcolor{blue}{0.170} & 0.227 & 0.194 & 0.205 & 0.214 & 0.636 \\ \hline

chatgpt-explicit & 0.164 & \textcolor{blue}{0.163} & 0.163 & 0.214 & \textbf{\textcolor{blue}{0.210}} & \textcolor{blue}{0.211} & \textcolor{blue}{0.218} & \textcolor{blue}{0.640}\\ \hline


chatgpt-hybrid & \textcolor{blue}{0.214} & 0.086 & 0.113 & \textcolor{blue}{0.307} & 0.126 & 0.174 & 0.116 & 0.626 \\ \hline

\multicolumn{9}{|c|}{\textbf{Davinci (chunk size 1024)}} 
\\ \hline

davinci-tldr  & \textbf{\textcolor{blue}{0.222}} & 0.143 & 0.169 & 0.260 & 0.166 & 0.197 & 0.207  & 0.641 \\ \hline

davinci-summ  & 0.196 & \textbf{\textcolor{blue}{0.181}} & \textbf{\textcolor{blue}{0.187}} & 0.223 & \textcolor{blue}{0.206} & \textbf{\textcolor{blue}{0.213}} & \textbf{\textcolor{blue}{0.219}}  & 0.638 \\ \hline

davinci-explicit & 0.175 & 0.159 & 0.166 & 0.214 & 0.195 & 0.203 & 0.209  & \textbf{\textcolor{blue}{0.643}}\\ \hline


davinci-hybrid & 0.172 & 0.144 & 0.154 & \textcolor{blue}{0.295} & 0.113 & 0.154  & 0.121  & 0.602 \\ \hline

\multicolumn{9}{|c|}{\textbf{Chatgpt-16k (chunk size 8192)}} 
\\ \hline
chatgpt-16k-long & 0.153 & 0.144 & 0.148 & 0.213 & 0.199 & 0.205 & 0.183 & 0.619\\ \hline

\end{tabular}
}
\caption{Summarization performances of general-purpose LLMs on the UK-Abs dataset for different prompts.
The value shown in blue is the best value in every summarization family for every metric. The value shown in bold is the highest value for every metric.}
\label{tab:general-metrics-ukabs-rouge}
\end{table*}

\begin{table*}[tb]
\small
\centering
\scalebox{0.9}
{
\begin{tabular}{l|lll|lll|ll}
\hline
\textbf{Model} & \textbf{R2-P} & \textbf{R2-R} & \textbf{R2-F1} & \textbf{RL-P} & \textbf{RL-R} & \textbf{RL-F1} & \textbf{ME} & \textbf{BS}  
\\
\hline
\multicolumn{9}{|c|}{\textbf{General domain LLMs}} \\ \hline
gpt4-explicit & 0.184 & 0.146 & 0.168 & \textcolor{blue}{0.228*} & 0.179 & 0.195 & 0.206 & 0.623 \\ \hline

chatgpt-explicit & 0.164 & 0.163 & 0.163 & 0.214 & \textbf{\textcolor{blue}{0.210*}} & 0.211* & 0.218 & \textbf{\textcolor{blue}{0.640}}

\\ \hline

llama-explicit & 0.165 & 0.129 & 0.148 & 0.206 & 0.173 & 0.181 & 0.200 & 0.629\\ \hline
davinci-summ & \textcolor{blue}{0.196} & \textbf{\textcolor{blue}{0.181}} & \textbf{\textcolor{blue}{0.187}} & 0.223* & 0.206* & \textbf{\textcolor{blue}{0.213*}} &  \textbf{\textcolor{blue}{0.219}} & 0.638 
\\ \hline





\multicolumn{9}{|c|}{\textbf{Legal domain-specific abstractive models}} \\ \hline

LegPegasus & \textbf{\textcolor{blue}{0.249*}} & 0.085 & 0.101 & \textbf{\textcolor{blue}{\textcolor{blue}{0.248*}}} & 0.113 & 0.136 & \textcolor{blue}{0.195} & \textcolor{blue}{0.635} \\ \hline

LegLED & 0.108 & \textcolor{blue}{0.107} & \textcolor{blue}{0.107} & 0.151 & \textcolor{blue}{0.149} & \textcolor{blue}{0.150} & 0.173  & 0.601\\ \hline



\multicolumn{9}{|c|}{\textbf{Extractive models}} 
\\ \hline

HipoRank & 0.161 & 0.147 & 0.155 & 0.165 & 0.140 & 0.161 & 0.204  & 0.602 \\ \hline

PACSUM & 0.152 & 0.140 & 0.150 & 0.158 & 0.138 & 0.141 & 0.202  & 0.606 \\ \hline

SummaRunner & 0.148 & 0.144 & 0.146 & 0.153 & 0.149 & 0.151 & 0.209  & 0.617 \\ \hline

CaseSummarizer & \textcolor{blue}{0.173} & \textcolor{blue}{0.164} & \textcolor{blue}{0.168} & \textcolor{blue}{0.174} & \textcolor{blue}{0.165} & \textcolor{blue}{0.169} & 0.204  & \textcolor{blue}{0.635}\\ \hline



BertSum & 0.158 & 0.156 & 0.157 & 0.158 & 0.155 & 0.156 & \textcolor{blue}{0.217}  & 0.624 \\ \hline

\end{tabular}
}
\caption{Comparing performances of the  best performing general-purpose LLM variants, legal domain-specific abstractive models, and extractive summarization models on the UK-Abs dataset.
The value shown in blue is the best value in every summarization family for every metric. The value shown in bold is the highest value for every metric.
Entries with an asterisk (*) indicate a value that is statistically significantly higher in terms of the student T-test at a 95\% confidence interval than the best value achieved by an extractive summarization model for the same metric.}
\label{tab:general-metrics-ukabs-first-extractive-abstractive}
\end{table*}

\vspace{2mm}
\noindent {\bf Comparing general-domain LLMs with domain-specific abstractive summarizers and extractive models:}
Table~\ref{tab:general-metrics-ukabs-first-extractive-abstractive} shows the ROUGE scores, METEOR, and BERTScore metrics for some of the best-performing general-domain LLM versions, legal domain-specific abstractive models and extractive summarization models on the UK-Abs dataset.

Amongst the extractive summarization models, CaseSummarizer has performed better than other extractive methods across most metrics.
Among the legal domain-specific abstractive models, 
LegLED achieves slightly higher ROUGE-2 and ROUGE-L Recall and F1 scores, while LegPegasus achieves higher METEOR and BERTScore values, and ROUGE-2 and ROUGE-L Precision scores.

Across all methods, the highest Rouge-2 recall, Rouge-2 F1 score, Rouge-L F1 score, and the highest METEOR score are achieved by davinci-summ. 
We see that the general-domain LLMs (davinci-summ in particular) perform better than the extractive models for the UK-Abs dataset.

The highest Rouge-2 Precision and Rouge-L Precision are obtained for LegPegasus, whereas LegLED achieves higher Rouge-2 Recall and Rouge-L Recall scores. 
We observe that the summaries generated by LegPegasus are generally shorter than the summaries generated by LegLED for the UK-Abs dataset. Specifically, the average length of summaries generated by LegPegasus is 820.77 words, while that of LegLED is 890.63 words. This difference in length of the summaries plays a role in higher ROUGE precision scores for LegPegasus and higher ROUGE Recall scores for LegLED. 

\begin{table}[tb]
\centering
\begin{tabular}{llll}
\hline
\textbf{Model} & \textbf{SummaC} & \textbf{NEPrec} & \textbf{NumPrec} \\ \hline
\hline

\multicolumn{4}{|c|}{\textbf{Llama2-70b}} \\ \hline

llama-tldr & 0.568 & \textcolor{blue}{0.917} & 0.925 \\ \hline

llama-summ & 0.558 & 0.906 & 0.921 \\ \hline

llama-hybrid & 0.604 & 0.887 & \textcolor{blue}{0.947}  \\ \hline


llama-explicit & \textcolor{blue}{0.606} & 0.875 & 0.936  \\ \hline

\multicolumn{4}{|c|}{\textbf{GPT4-Turbo}} \\ \hline

gpt4-tldr & 0.595 & \textbf{\textcolor{blue}{0.956}} & 0.983 \\ \hline

gpt4-summ & 0.586 & 0.937 & 0.955 \\ \hline

gpt4-hybrid & \textcolor{blue}{0.641} & 0.927 & \textbf{\textcolor{blue}{0.984}}  \\ \hline


gpt4-explicit & 0.636 & 0.927 & 0.979  \\ \hline

\multicolumn{4}{|c|}{\textbf{Chatgpt}} \\ \hline

chatgpt-tldr & 0.588 & \textcolor{blue}{0.934} & \textcolor{blue}{0.965} \\ \hline

chatgpt-summ & 0.570 & 0.921 & 0.933 \\ \hline

chatgpt-hybrid & \textcolor{blue}{0.630} & 0.902 & 0.963  \\ \hline


chatgpt-explicit & 0.622 & 0.900 & 0.952  \\ \hline

\multicolumn{4}{|c|}{\textbf{Davinci}} \\ \hline

davinci-tldr & 0.623 & 0.833 & 0.912  \\ \hline

davinci-summ & 0.634 & \textcolor{blue}{0.926} & \textcolor{blue}{0.977}  \\ \hline

davinci-hybrid & 0.624 & 0.903 & 0.956  \\ \hline

davinci-explicit & \textbf{\textcolor{blue}{0.677}} & 0.887 & 0.963  \\ \hline

\multicolumn{4}{|c|}{\textbf{Chatgpt-16k}} \\ \hline

chatgpt-16k-long &  0.644 & \textbf{\textcolor{blue}{0.956}} & 0.981 \\ \hline


\multicolumn{4}{|c|}{\textbf{Legal domain-specific abstractive models}} \\ \hline
LegLED & 0.628 & 0.874 & 0.914 \\ \hline
LegPegasus & \textcolor{blue}{0.649} & \textcolor{blue}{0.907} & \textcolor{blue}{0.959} \\ \hline

\end{tabular}

\caption{Consistency metrics for summaries generated by various models on UK-Abs dataset. The value shown in blue is the best value in every summarization family for every metric. The value shown in bold is the overall highest value for every metric.
}
\label{tab:consistency-metrics-ukabs-gen}
\end{table}

\vspace{2mm}
\noindent {\bf Consistency of generated summaries:}
Table~\ref{tab:consistency-metrics-ukabs-gen} shows the consistency metrics for general-purpose LLMs and the domain-specific abstractive models on the UK-Abs dataset. The value shown in blue is the best value in every summarization family for every metric. The value shown in blue-bold is the highest value for every metric.
We found these metric values to be 1.0 for the extractive summaries, hence we do not show them in the table.\footnote{Note that a recent study has shown that even extractive summaries can have different types of inconsistencies~\cite{zhang-etal-2023-extractive}. However, the metrics we applied gave value 1.0 for the extractive summaries.}

Among the chatgpt variants, 
chatgpt-tldr gets the highest NEPrec and NumPrec scores, while chatgpt-hybrid gets the highest SummaC score.
Among the gpt4 variants, 
gpt4-tldr gets the highest NEPrec scores, while gpt4-hybrid gets the highest SummaC and NumPrec scores. Among the llama variants, 
llama-tldr gets the highest NEPrec scores, while llama-hybrid gets the highest NumPrec score and llama-explicit gets the highest SummaC score.
Whereas, among the davinci variants, davinci-summ gets the highest 
NEPrec and NumPrec scores, and the highest SummaC score is obtained by Davinci-explicit (which is the overall highest score for SummaC).
The overall highest values for NEPrec and Numprec scores are achieved by GPT4-Turbo (NEPrec jointly highest with chatgpt-16k-long).

From Table~\ref{tab:consistency-metrics-ukabs-gen}, it is evident that there are instances of hallucination / inconsistency in the summaries generated by LLMs and other generative summarization models. 
We will see some actual examples of such inconsistencies in the next section.

\subsection{Summarization results on IN-Abs dataset}

We now study the performances of various summarization models on the IN-Abs dataset.
As we did in the previous section for the UK-Abs dataset, here also we first compare the performances of LLM-based summarization using different prompts (that were stated in Section~\ref{sub:llms-for-summarization}), and select the best LLM variants for summarization of this dataset. 
Then we compare the best LLM variants with other extractive and abstractive summarization models.

\begin{table}[tb]
\small
\centering
\scalebox{0.9}
{
\begin{tabular}{l|lll|lll|llr}
\hline
\textbf{Model} & \textbf{R2-P} & \textbf{R2-R} & \textbf{R2-F1} & \textbf{RL-P} & \textbf{RL-R} & \textbf{RL-F1}  & \textbf{ME} & \textbf{BS} 
\\
\hline

\multicolumn{9}{|c|}{\textbf{Llama2-70b(chunk size 1024)}} 
\\ \hline

llama-tldr & \textcolor{blue}{0.243} & 0.113 & 0.148 & \textcolor{blue}{0.307} & 0.147 & 0.184 & 0.142  & \textcolor{blue}{0.594} \\ \hline

llama-summ & 0.194 & \textcolor{blue}{0.147} & \textcolor{blue}{0.162} & 0.241 & 0.181 & 0.185 & \textcolor{blue}{0.173}  & 0.593 \\ \hline

llama-explicit & 0.209 & 0.141 & 0.157 & 0.262 & \textcolor{blue}{0.182} & \textcolor{blue}{0.206}  & 0.158 & 0.591 \\ \hline

llama-hybrid & 0.161 & 0.122 & 0.131 & 0.246 & 0.147 & 0.165  & 0.129  & 0.581  \\ \hline

\multicolumn{9}{|c|}{\textbf{GPT4-Turbo }} 
\\ \hline

gpt4-tldr & \textbf{\textcolor{blue}{0.256}} & 0.126 & 0.159 & \textcolor{blue}{0.316} & 0.162 & 0.207 & 0.159  & 0.602 \\ \hline

gpt4-summ & 0.216 & \textcolor{blue}{0.158} & \textcolor{blue}{0.174} & 0.253 & 0.192 & 0.206 & \textcolor{blue}{0.183}  & \textcolor{blue}{0.618} \\ \hline

gpt4-explicit & 0.216 & 0.156 & 0.173 & 0.283 & \textcolor{blue}{0.209} & \textcolor{blue}{0.224}  & 0.172 & 0.604 \\ \hline

gpt4-hybrid & 0.183 & 0.127 & 0.141 & 0.263 & 0.155 & 0.175  & 0.149  & 0.601  \\ \hline

\multicolumn{9}{|c|}{\textbf{Chatgpt (chunk size 1024)}} 
\\ \hline

chatgpt-tldr & \textcolor{blue}{0.238} & 0.140 & 0.173 & \textcolor{blue}{0.298} & 0.175 & 0.204 & 0.167  & 0.609 \\ \hline

chatgpt-summ & 0.199 & \textcolor{blue}{0.177} & \textcolor{blue}{0.186} & 0.232 & 0.209 & 0.214 & \textbf{\textcolor{blue}{0.193}}  & \textbf{\textcolor{blue}{0.624}} \\ \hline

chatgpt-explicit & 0.205 & 0.174 & 0.175 & 0.267 & \textbf{\textcolor{blue}{0.224}} & \textbf{\textcolor{blue}{0.235}}  & 0.186 & 0.605 \\ \hline

chatgpt-hybrid & 0.195 & 0.134 & 0.144 & 0.257 & 0.173 & 0.188  & 0.155  & 0.616  \\ \hline


\multicolumn{9}{|c|}{\textbf{Davinci (chunk size 1024)}} 
\\ \hline
davinci-tldr & \textcolor{blue}{0.233} & 0.125 & 0.156 & \textcolor{blue}{0.284} & 0.152 & 0.190 & 0.141  & 0.604 \\ \hline

davinci-summ & 0.220 & \textbf{\textcolor{blue}{0.179}} & \textbf{\textcolor{blue}{0.195}} & 0.251 & \textcolor{blue}{0.205} & \textcolor{blue}{0.223} & \textcolor{blue}{0.191}  & \textcolor{blue}{0.624} \\ \hline

davinci-explicit & 0.169 & 0.125 & 0.140 & 0.245 & 0.188 & 0.209 &  0.170  & 0.616  \\ \hline

davinci-hybrid & 0.201 & 0.121 & 0.145 & 0.236 & 0.180 & 0.196  & 0.168  & 0.612 \\ \hline


\multicolumn{9}{|c|}{\textbf{Chatgpt-16k (chunk size 8192)}} 
\\ \hline
chatgpt-16k-long & 0.231 & 0.127 & 0.149 & \textbf{\textcolor{blue}{0.319}} & 0.176
 & 0.206 & 0.147  & 0.615  \\ \hline









\end{tabular}}

\caption{ROUGE, METEOR, and BERTScore metrics for general-domain LLMs on the IN-Abs dataset, for different prompts. 
The values shown in blue are the best values in every summarization family for every metric. The values shown in blue-bold are the highest values for every metric across all methods.
}
\label{tab:rouge-ukabs-prompting}
\end{table}

\vspace{2mm}
\noindent {\bf Comparing LLM-based summarization variants:}
Table~\ref{tab:rouge-ukabs-prompting} compares the performance of general-domain LLMs on the IN-Abs dataset, for different prompts. 
The best value for a metric within a particular family of summarization models is shown in blue, and the overall best value for every metric is shown in blue-bold. 

Amongst the chatgpt models, chatgpt-summ achieves the best value for most metrics, including METEOR and BERTScore, although chatgpt-explicit achieves the best ROUGE-L Recall and F1 scores. 
Amongst the gpt4 models, gpt4-summ achieves the best value for most metrics, including METEOR and BERTScore, although gpt4-explicit achieves the best ROUGE-L Recall and F1 scores. Among the llama models, different variants get the highest score for different metrics. 
Among the davinci models, davinci-summ achieves the highest values for most metrics. 

We observe that the `tldr' variants have higher Rouge-2 precision and Rouge-L precision across all the model variants. 
Whereas, the `summ' and `explicit' variants have higher Rouge-2 and Rouge-L recall and F1 scores across all the model variants.

An interesting observation is that the summaries produced by GPT-4 Turbo are shorter than those generated by Turbo-GPT-3.5 (Chatgpt). As a result, the summaries from Chatgpt exhibit higher ROUGE Recall and F1 scores, whereas the GPT-4 summaries achieve higher ROUGE precision scores.

We now compare some of the best performing LLM variants with the other types of summarization models.

\begin{table*}[tb]
\small
\centering
\scalebox{0.9}
{
\begin{tabular}{l|lll|lll|ll}
\hline
\textbf{Model} & \textbf{R2-P} & \textbf{R2-R} & \textbf{R2-F1} & \textbf{RL-P} & \textbf{RL-R} & \textbf{RL-F1} & \textbf{ME} & \textbf{BS}  
\\
\hline

\multicolumn{9}{|c|}{\textbf{General domain LLMs}} \\ \hline

gpt4-summ & 0.216 & 0.153 & 0.174 & \textcolor{blue}{0.253} & 0.192 & 0.206 & 0.183  & 0.618 \\ \hline

chatgpt-summ & 0.199 & 0.177 & 0.186 & 0.232 & \textbf{\textcolor{blue}{0.209}} & 0.214 & \textcolor{blue}{0.193}  & \textbf{\textcolor{blue}{0.624}} \\ \hline

llama-summ & 0.194 & 0.147 & 0.162 & 0.241 & 0.181 & 0.185 & 0.173  & 0.593 \\ \hline


davinci-summ & \textcolor{blue}{0.220} & \textcolor{blue}{0.179} & \textcolor{blue}{0.195} & 0.251 & 0.205 & \textcolor{blue}{\textbf{0.223}} & 0.191  & \textbf{\textcolor{blue}{0.624}} \\ \hline










\multicolumn{9}{|c|}{\textbf{Legal domain-specific abstractive models}} \\ \hline

LegPegasus & \textcolor{blue}{0.196} & \textcolor{blue}{0.120} & \textcolor{blue}{0.133} & \textbf{\textcolor{blue}{0.263*}} & \textcolor{blue}{0.154} & \textcolor{blue}{0.172} & \textcolor{blue}{0.194}  & \textcolor{blue}{0.594} \\ \hline

LegLED & 0.111 & 0.107 & 0.108 & 0.150 & 0.146 & 0.147 & 0.142  & 0.590 \\ \hline






\multicolumn{9}{|c|}{\textbf{Extractive models}} 
\\ \hline

PACSUM  & 0.228 & 0.207 & 0.216 & 0.229 & 0.185 & 0.202 &  0.200  & 0.603 \\ \hline

HipoRank  & 0.221 & 0.204 & 0.207 & 0.214 & 0.184 & 0.194 &  0.194  & 0.602 \\ \hline

SummaRunner & 0.227 & 0.210 & 0.218 & 0.198 & 0.182 & 0.189 & \textbf{\textcolor{blue}{0.203}}  & \textcolor{blue}{0.621} \\ \hline

CaseSummarizer & \textbf{\textcolor{blue}{0.251}} & \textbf{\textcolor{blue}{0.226}} & \textbf{\textcolor{blue}{0.238}} & \textcolor{blue}{0.231} & \textcolor{blue}{0.208} & \textcolor{blue}{0.219} & 0.194  & 0.614\\ \hline

BertSum & 0.248 & 0.218 & 0.232 & 0.226 & 0.196 & 0.209 &  0.202  & 0.615 \\ \hline

\end{tabular}}

\caption{ROUGE, METEOR, BERTScore metrics for best performing general-domain LLMs, abstractive and extractive summarization models on the IN-Abs dataset.
The value shown in blue is the best value in every summarization family for every metric. The value shown in blue-bold is the highest value for every metric.
Entries with an asterisk (*) indicate a value that is statistically significantly higher in terms of the student T-test at a 95\% confidence interval than the best value achieved by an extractive summarization model.
}
\label{tab:in-abs_extractive_abstrative-first}
\end{table*}

\begin{table}[tb]
\centering
\begin{tabular}{llll}
\hline
\textbf{Model} & \textbf{SummaC} & \textbf{NEPrec} & \textbf{NumPrec} \\ \hline

\multicolumn{4}{|c|}{\textbf{Llama2-70b}} 
\\ \hline

llama-tldr & 0.559 & 0.846 & 0.927 \\ \hline

llama-summ & 0.558 & 0.883 & 0.921  \\ \hline

llama-hybrid & \textcolor{blue}{0.605} & \textcolor{blue}{0.904} & 0.904  \\ \hline


llama-explicit & 0.579 & 0.884 & \textcolor{blue}{0.928}   \\ \hline

\multicolumn{4}{|c|}{\textbf{GPT4-Turbo}} 
\\ \hline

gpt4-tldr & 0.581 & 0.905 & 0.966 \\ \hline

gpt4-summ & 0.596 & \textcolor{blue}{0.942} & 0.978  \\ \hline

gpt4-hybrid & \textcolor{blue}{0.652} & 0.927 & 0.973  \\ \hline


gpt4-explicit & 0.631 & 0.909 & \textbf{\textcolor{blue}{0.986}}   \\ \hline

\multicolumn{4}{|c|}{\textbf{Chatgpt}} 
\\ \hline

chatgpt-tldr & 0.570 & 0.851 & 0.943 \\ \hline

chatgpt-summ & 0.573 & 0.901 & 0.956  \\ \hline

chatgpt-hybrid & \textcolor{blue}{0.635} & \textcolor{blue}{0.907} & 0.963  \\ \hline


chatgpt-explicit & 0.608 & 0.886 & \textcolor{blue}{0.974}   \\ \hline

\multicolumn{4}{|c|}{\textbf{Davinci}} 
\\ \hline

davinci-summ & 0.635 & 0.895 & 0.932  \\ \hline

davinci-tldr & 0.608 & 0.833 & 0.912   \\ \hline


davinci-explicit & \textbf{\textcolor{blue}{0.694}} & 0.831 & \textcolor{blue}{0.957} \\ \hline

davinci-hybrid & 0.622 & \textcolor{blue}{0.904} & 0.956  \\ \hline

\multicolumn{4}{|c|}{\textbf{Chatgpt-16k}} 
\\ \hline

chatgpt-16k-long & 0.633 & \textbf{\textcolor{blue}{0.944}} & 0.975 \\ \hline


\multicolumn{4}{|c|}{\textbf{Legal Domain-specific abstractive models}} 
\\ \hline

LegPegasus & 0.633 & \textcolor{blue}{0.842} & \textcolor{blue}{0.948} \\ \hline

LegLED & \textcolor{blue}{0.656} & 0.719 & 0.819 \\ \hline

\end{tabular}

\caption{Consistency metrics for different abstractive summarization models on IN-Abs dataset. The value shown in blue is the best value in every summarization family for every metric. The value shown in bold is the highest value for every metric.
}
\label{tab:consistency-metrics-inabs-prompting}
\end{table}

\vspace{2mm}
\noindent {\bf Comparing general-domain LLMs with domain-specific abstractive summarizers and extractive models:}
Table~\ref{tab:in-abs_extractive_abstrative-first} shows the ROUGE, METEOR, BERTScore metrics for best performing general-domain LLMs, abstractive and extractive summarization models on the IN-Abs dataset.

Among the extractive models, CaseSummarizer achieves the highest metric values for most metrics, especially the ROUGE-based metrics. In fact, for most ROUGE-based metrics, CaseSummarizer achieves the overall highest values for all metrics. 
Among the legal domain-specific abstractive models, LegPegasus performs better than LegLED across all metrics. 
Among the LLMs, chatgpt-summ achieves higher ROUGE-L Recall and BertScore values than the extractive models. 
Thus for the IN-Abs dataset, it is seen that the abstractive models (including LLMs) perform at par with the best extractive models. 


\vspace{2mm}
\noindent {\bf Consistency of generated summaries:}
Table~\ref{tab:consistency-metrics-inabs-prompting} shows the consistency metrics for summaries generated by all abstractive models on the IN-Abs dataset.
The value shown in blue is the best value in every summarization family for every metric. The value shown in blue-bold is the overall highest value for every metric.
We found these metric values to be 1.0 for the extractive summaries, hence we do not show them in the table.

Among the chatgpt variants, 
chatgpt-hybrid gets the highest SummaC and NEPrec scores, while chatgpt-explicit gets the highest NumPrec score.
Among the gpt4 variants, gpt4-hybrid gets the highest SummaC scores, while gpt4-explicit gets the highest NumPrec score (which is the overall highest score for NumPrec) and gpt4-summ gets the highest NEPrec scores. Among the llama2-70b variants, llama-hybrid gets the highest SummaC and NEPrec scores, while llama-explicit gets the highest NumPrec score.
Among the davinci variants, davinci-explicit gets the highest 
SummaC (highest overall) and NumPrec scores, and the highest NEPrec score is obtained by davinci-hybrid.

Again, it is evident that there are instances of hallucination / inconsistency in the summaries generated by LLMs and other generative summarization models. 
We will see some actual examples of such inconsistencies in the next section.

\section{Examples of inconsistencies and hallucinations in abstractive summaries}
\label{sec:inconsistency_examples}


\begin{table*}[tb]
\centering
\small
\begin{tabular}{|p{0.01\textwidth}|p{0.15\textwidth}|p{0.35\textwidth}|p{0.35\textwidth}|}
\hline
\textbf{id} & \textbf{Model} & \textbf{Extract from summary showing error} & \textbf{Explanation of error}  \\ \hline

1
&
chatgpt-summ
& 
... \textcolor{red}{27,000} in currency notes on the pretext that he wanted to pay the balance of \textcolor{red}{Rs 26,500} due to the accused under the agreement alleged to have been entered into between them ...
&
There is no mention of Rs. 26,500 or Rs. 27,000 in the main document. Both these amounts are hallucinated. The amount stated in the original document is Rs. 29,500.
\\ \hline

2
&
chatgpt-summ
&
On these materials, the learned Magistrate held that the appellant had committed an offence under \textcolor{red}{Section 387 of the Indian Penal Code} and convicted him accordingly. 
&
Section 387 of the Indian Penal Code is not mentioned in the main document. This Section 387 is hallucinated. \\ \hline

3
&
davinci-tldr 
& 
The Supreme Court accepted the argument raised on behalf of the
& 
Incomplete line
\\ \hline





4
&
LegLED
& 
On February 1, \textcolor{red}{2019}, the Honorable \textcolor{red}{Chandrasekhar A. Lama} of the \textcolor{red}{Presidency of the United Kingdom} entered a final judgment against W.H. King......
& 
The name of the judge according to the original judgement is  ``Aiyar, N. Chandrasekhara'' which is written wrongly in the summary. The judge clearly does not belong to the Presidency of United Kingdom, since this is a case in the Indian Supreme Court. Also, the year 2019 does not occur in the judgement.
\\ \hline

\end{tabular}
\caption{
Examples of errors in abstractive summaries generated by different summarization models for the Indian Court judgement available at \url{https://indiankanoon.org/doc/1801104/}. The errors in the summaries are marked in red, and explained in the last column. 
}
\label{tab:inconsistency-example-1}
\end{table*}


\begin{table*}[tb]
\centering
\small
\begin{tabular}{|p{0.01\textwidth}|p{0.10\textwidth}|p{0.40\textwidth}|p{0.35\textwidth}|}
\hline
\textbf{id} & \textbf{Model} & \textbf{Extract from summary showing error} & \textbf{Explanation of error}  \\ \hline
\hline
1
&
LegLED
&
On March 15, 2019, the \textcolor{red}{Honorable M.L. Mehta} of the \textcolor{red}{U.S. District Court} for the District of Ludhiana entered a final judgment against the defendants in the \textcolor{red}{Ludhiana, Ludhiana, Ludhiana, Ludhiana, and Ludhiana} Revision Petition No. 17 R/55 of 1955.
&
There is an advocate named ``K. L. Mehta'' but no judge named ``M. L. Mehta'' in the original judgement. The word ``Ludhiana'' has been repeated multiple times. Also there is no mention of U.S. District Court in the judgement. \\ \hline



2
&
chatgpt-summ 
& 
... as their application for the allotment of Raikot land was rejected by the Deputy Custodian General on the ground that his jurisdiction to revise the order has been taken away by virtue of the provisions of, \textcolor{red}{(44 of 1954)} and the notification issued thereunder on March 24, 1955.
& 
The original document states ``Displaced Persons (Compensation and Rehabilitation) Act, 1954 (44 of 1954)'' but the name of the Act is not mentioned in the summary; a critical detail is omitted.

\\ \hline

3
&
LegLED 
& 
According to the \textcolor{red}{SEC's complaint filed in the U.S. District Court for the Southern District of New York}, \textcolor{red}{from March 24, 1955 to March 24, 1955}, the Raikot lands were allotted to the appellants......
& 
The summary talks about ``U.S. District Court, SEC'', and ``Southern District of New York'' which are completely unrelated to this case, since the case is in the Indian Supreme Court.
Also ``from March 24, 1955 to March 24,1955'' is meaningless.
\\ \hline





\end{tabular}
\caption{
Examples of errors in abstractive summaries generated by different summarization models for the Indian Court judgement available at \url{https://indiankanoon.org/doc/302141/}. The errors in the summaries are marked in red, and explained in the last column. 
}
\label{tab:inconsistency-example-2}
\end{table*}


\begin{table*}[tb]
\centering
\small
\begin{tabular}{|p{0.01\textwidth}|p{0.15\textwidth}|p{0.35\textwidth}|p{0.35\textwidth}|}
\hline
\textbf{id} & \textbf{Model} & \textbf{Extract from summary showing error} & \textbf{Explanation of error}  \\ \hline
\hline
1
&
Davinci-summ 
& 
The Supreme Court has allowed an appeal by a woman against the Serious Organised Crime Agency (SOCA) over the agency’s attempt to recover \textcolor{red}{£4.5m} from her.
& 
There is no mention of £4.5m in the main document. Rather the main document mentions ``US \$2.5m''.
Rather there is a mention of Clause 4.5 in the settlement deed. The model has confused between the clause 4.5 and the amount \$2.5m.
\\ \hline

2
&
LegLED
&
The Honorable Lord Neuberger has agreed to the entry of a final judgment that permanently enjoins him from future violations of the antifraud provisions of \textcolor{red}{Section 17(a) of the Securities Act of 1933 (``Securities Act'')} ....
&
The Section 17(a) of the Securities Act of 1933 is hallucinated. No such Act is present in the main judgement. This is, in fact, an US Act whereas the input case is of the UK Supreme Court.
\\ \hline

\end{tabular}
\caption{
Examples of errors in abstractive summaries generated by different summarization models for the UK Court judgement available at \url{https://www.supremecourt.uk/cases/docs/uksc-2011-0196-judgment.pdf}. The errors in the summaries are marked in red, and explained in the last column. 
}
\label{tab:inconsistency-example-3}
\end{table*}


\begin{table*}[tb]
\centering
\small
\begin{tabular}{|p{0.01\textwidth}|p{0.15\textwidth}|p{0.35\textwidth}|p{0.35\textwidth}|}
\hline
\textbf{id} & \textbf{Model} & \textbf{Extract from summary showing error} & \textbf{Explanation of error}  \\ \hline
\hline


1
&
Chatgpt-tldr
&
The closest any international instrument has come to provide for such a general immunity is article 11.2(b) of the \textcolor{red}{United Nations Convention on Jur The} European Court of Human Rights has only considered article 11 in cases ......
& 
The phrase marked in red (actually a statute name) is incomplete and has been merged with the next sentence. The entire phrase as stated in the original judgement is ``United Nations Convention on Jurisdictional Immunities of States and their Property (2004)''.
\\ \hline

2
&
Davinci-summ
&
The case was brought by two women, one Moroccan and \textcolor{red}{one Tunisian}, who sued the embassies of Sudan and Libya respectively .....
&
The original judgement does not have the country `Tunisia' anywhere in the judgement. This information is hallucinated. The judgement talks about two women both of whom are from Morocco.
\\ \hline

3
&
LegLED
&
The \textcolor{red}{U.S. District Court for the Western District of New York today entered a final judgement in favor of the Kingdom of Saudi Arabia} in a case involving claims to state immunity.
&
The U.S. District Court is nowhere stated in the original legal judgement. So this entire information is hallucinated.
\\ \hline
\end{tabular}
\caption{
Examples of errors in abstractive summaries generated by different summarization models for the UK Court judgement available at \url{https://www.supremecourt.uk/cases/docs/uksc-2015-0063-judgment.pdf}. The errors in the summaries are marked in red and explained in the last column. 
}
\label{tab:inconsistency-example-4}
\end{table*}

In the context of generative models such as LLMs and abstractive summarizers, \textit{hallucination} refers to the generation of text that is not based on real or accurate information but \textit{appears to be} coherent and contextually relevant~\cite{hallucination-survey}. 
These models generate responses/outputs by predicting the next word or sequence of words based on patterns learned from vast amounts of training data. Occasionally, these models can generate responses that are factually incorrect, nonsensical, or entirely made up, resembling hallucinations in human perception. 

From the relatively low values of the SummaC metric that we observed in the previous section, it appears that the LLMs and legal abstractive summarizers are likely to hallucinate parts of the summary. 
To check if this is the case, we manually observed several (document, summary) pairs in the two datasets and found some inconsistencies / hallucinations in almost all of the summaries that we manually checked. 
In particular, we observed those sentences that obtained relatively low SummaC scores, and those sentences that contained numbers and named entities that could \textit{not} be matched with the original documents (while computing NERPrec and NumPrec). We also observed the relevant parts in the main documents to understand the errors/inconsistencies.

We present some examples of errors/inconsistencies in this section.
Table~\ref{tab:inconsistency-example-1} and Table~\ref{tab:inconsistency-example-2} show examples of such errors in the summaries of two Indian Supreme Court cases, while Table~\ref{tab:inconsistency-example-3} and Table~\ref{tab:inconsistency-example-4} show examples of errors in summaries of two cases of the UK Supreme Court. 
In all tables, the summarization model that committed the error is stated, an extract from the summary showing the error is given, and the error is explained briefly.

In Table~\ref{tab:inconsistency-example-1}, the first example shows two wrong monetary values that are stated in the summary generated by chatgpt-summ, but are \textit{not} present in the main judgement. The second example shows a hallucinated statute included in the summary generated by chatgpt-summ, that is not present in the main judgement. 
Other examples show incomplete lines in the summary, and names of persons wrongly stated in the summary. 
Table~\ref{tab:inconsistency-example-2} also shows examples where the name of the judge is stated wrongly in the summary generated by LegLED, the name of an important statute is omitted from the summary generated by chatgpt-summ, and where LegLED has hallucinated the name of a court in the USA while summarizing a judgement of the Indian Supreme Court.


Table~\ref{tab:inconsistency-example-3} and Table~\ref{tab:inconsistency-example-4} show examples from two UK Supreme Court case judgement summaries. 
In Table~\ref{tab:inconsistency-example-3}, a wrong monetary value in mentioned in the summary in the first example (actually, Davinci-summ possibly confused the amount ``\$2.5m'' with ``Clause 4.5'', both of which appear in the judgement, and wrongly generated ``4.5m'' as the amount).
The second example shows a statute that is included in the summary (``Section 17(a) of Securities Act of 1933'') but which is absent in the main judgement. So this information is hallucinated. 
Table~\ref{tab:inconsistency-example-4} also shows examples of hallucinated country names and court names, as well as instances where a statute name is only incompletely stated in the summary. 


We noticed one strange type of error particularly in summaries generated by LegLED -- even when the model is summarizing Indian or UK case judgements, names of U.S. Courts and names of U.S. statutes come up in the summaries, which are not at all related to the input document (several examples are given in the tables). 
Such hallucinations are probably due to the fact that LegLED has been trained on US legal document-summary pairs, and the model has a tendency of generating US court / statute names that it has seen during training.

Along with the more serious errors that we have shown -- such as names and monetary values being reported wrongly, court names and statute names being hallucinated, incomplete statute names -- the abstractive summarizers and LLMs also frequently commit smaller errors such as two words being merged or two sentences being merged (a sentence beginning without the previous one ending). 

\section{Exploring ways to reduce inconsistencies in abstractive summaries}
\label{sec:reduce_inconsistency}

From the previous section, it is evident that hallucinations / inconsistencies are observed in the abstractive summaries of legal judgements. 
We now explore three different strategies for reducing hallucinations in abstractive summaries, one for abstractive summarizers (that allow fine-tuning) and two methods for LLMs (for which fine-tuning is not possible or prohibitively expensive).

\subsection{Domain-specific fine-tuning of abstractive models}

As stated earlier, the abstractive summarization models LegPegasus and LegLED are originally fine-tuned over legal (case judgement, summary) pairs from courts in the USA. 
These models allow further fine-tuning. Hence we check if domain-specific fine-tuning (i.e., fine-tuning over data from the target domain) of these models helps in improving consistency of the generated summaries.

To make these models more suitable for summarizing UK legal documents, we further fine-tune them on the UK-Abs training dataset containing 693 UK case judgements and their corresponding reference summaries. These models fine-tuned on the UK-Abs training dataset are referred to as \textbf{LegPegasus-UK} and \textbf{LegLED-UK}. 
Similary, to make these models more suitable for Indian case judgements, LegLED and LegPegasus models are further finetuned on the IN-Abs training dataset consisting of 7030 (case judgement, summary) pairs. We refer to these models fine-tuned over Indian data as \textbf{LegLED-IN} and \textbf{LegPegasus-IN}.

\begin{table}[tb]
\centering
 \scalebox{0.9}{
\begin{tabular}{lllllllll}
\hline
\textbf{Model} & \textbf{R2-P} & \textbf{R2-R} & \textbf{R2-F1} & \textbf{RL-P} & \textbf{RL-R} & \textbf{RL-F1} & \textbf{ME} & \textbf{BS}  \\ \hline


LegPegasus & 0.196 & 0.120 & 0.133 & 0.263 & 0.154 & 0.172 & 0.194  & 0.594 \\ \hline

LegPegasus-IN & \textbf{\textcolor{blue}{0.264}} & \textcolor{blue}{0.243} & \textcolor{blue}{0.251} & \textbf{\textcolor{blue}{0.281}} & \textcolor{blue}{0.262} & \textcolor{blue}{0.269} & \textcolor{blue}{0.196}  & \textbf{\textcolor{blue}{0.624}} \\ 

\hline \hline

LegLED & 0.111 & 0.107 & 0.108 & 0.150 & 0.146 & 0.147 & 0.142  & 0.590 \\ \hline

LegLED-IN & \textcolor{blue}{0.260} & \textbf{\textcolor{blue}{0.253}} & \textbf{\textcolor{blue}{0.255}} & \textcolor{blue}{0.276} & \textbf{\textcolor{blue}{0.269}} & \textbf{\textcolor{blue}{0.271}} &  \textbf{\textcolor{blue}{0.226}}  & \textbf{\textcolor{blue}{0.624}} \\ \hline
\end{tabular}}

\caption{Comparing the performance of abstractive summarization models over the IN-Abs test set, before and after domain-specific fine-tuning. The higher value for every metric is shown in blue, and the overall highest value for every metric is in blue-bold.
}
\label{tab:automated-metrics-inabs}
\end{table}

\begin{table}[tb]
\centering
 \scalebox{0.9}{
\begin{tabular}{llll}
\hline
\textbf{Model} & \textbf{SummaC} & \textbf{NEPrec} & \textbf{NumPrec} \\ \hline

LegPegasus & 0.633 & 0.842 & 0.948 \\ \hline

LegPegasus-IN & \textcolor{blue}{0.736} & \textbf{\textcolor{blue}{0.854}} & \textbf{\textcolor{blue}{0.995}} \\ 

\hline \hline

LegLED & 0.656 & 0.719 & 0.819 \\ \hline

LegLED-IN & \textbf{\textcolor{blue}{0.855}} & \textcolor{blue}{0.827} & \textcolor{blue}{0.976} \\ \hline

\end{tabular}}

\caption{Consistency metrics for abstractive summarization models over the IN-Abs dataset, before and after domain-specific  fine-tuning. The higher value for every metric is shown in blue, and the overall highest value for every metric is in blue-bold.
}
\label{tab:reducing_hallucinaion_inabs_consistency}
\end{table}

\begin{table}[tb]
\centering
\scalebox{0.9}{
\begin{tabular}{lllllllll}
\hline
\textbf{Model} & \textbf{R2-P} & \textbf{R2-R} & \textbf{R2-F1} & \textbf{RL-P} & \textbf{RL-R} & \textbf{RL-F1} & \textbf{ME} & \textbf{BS}  \\ \hline

LegPegasus & \textbf{\textcolor{blue}{0.249}} & 0.085 & 0.101 & \textbf{\textcolor{blue}{0.248}} & 0.113 & 0.136 & 0.195 & 0.635 \\ \hline

LegPegasus-UK & 0.183 & \textbf{\textcolor{blue}{0.173}} & \textbf{\textcolor{blue}{0.178}} & 0.217 & \textbf{\textcolor{blue}{0.205 }} & \textbf{\textcolor{blue}{0.211}} & \textcolor{blue}{0.200} & \textbf{\textcolor{blue}{0.640}} \\ 

\hline \hline

LegLED & 0.108 & 0.107 & 0.107 & 0.151 & 0.149 & 0.150 & 0.173 & 0.601 \\ \hline

LegLED-UK & \textcolor{blue}{0.164} & \textcolor{blue}{0.155} & \textcolor{blue}{0.159} & \textcolor{blue}{0.194} & \textcolor{blue}{0.182} & \textcolor{blue}{0.188} & \textbf{\textcolor{blue}{0.208}} & \textcolor{blue}{0.637} \\ \hline

\end{tabular}}

\caption{Comparing the performance of abstractive summarization models over the UK-Abs test set, before and after domain-specific fine-tuning.  The higher value for every metric is shown in blue, and the overall highest value for every metric is in blue-bold.
}
\label{tab:automated-metrics-ukabs2}
\end{table}

Table~\ref{tab:automated-metrics-inabs} shows the ROUGE, METEOR, and BERTScore for legal domain-specific abstractive models on the IN-Abs dataset before (LegPegasus and LegLED) and after (LegPegasus-IN and LegLED-IN) using fine-tuning. 
Table~\ref{tab:reducing_hallucinaion_inabs_consistency} shows the consistency metrics over IN-Abs for the same scenarios.
It is seen that both the match of the generated summaries with reference summaries (Table~\ref{tab:automated-metrics-inabs}) as well as the consistency of generated summaries (Table~\ref{tab:reducing_hallucinaion_inabs_consistency}) improve after fine-tuning over data from the target domain.

Similarly, Table~\ref{tab:automated-metrics-ukabs2} and Table~\ref{tab:reducing_hallucinaion_ukabs_consistency} compare the performance of the abstractive summarization models on the UK-Abs dataset, before and after fine-tuning over target domain data. 
We again see that performance as well as consistency of the generated summaries improves massively upon fine-tuning over the target domain data.



\begin{table}[tb]
\centering
\begin{tabular}{llll}
\hline
\textbf{Model} & \textbf{SummaC} & \textbf{NEPrec} & \textbf{NumPrec} \\ \hline


LegPegasus & 0.649 & \textbf{\textcolor{blue}{0.907}} & 0.959 \\ \hline

LegPegasus-UK & \textcolor{blue}{0.708} & \textbf{\textcolor{blue}{0.907}} & \textbf{\textcolor{blue}{0.992}} \\ 

\hline \hline

LegLED & 0.628 & 0.874 & 0.914 \\ \hline

LegLED-UK & \textbf{\textcolor{blue}{0.835}} & \textcolor{blue}{0.882} & \textcolor{blue}{0.989} \\ \hline

\end{tabular}

\caption{Consistency metrics for legal domain-specific abstractive models for UK-Abs dataset before and after domain-specific fine-tuning. The higher value for every metric is shown in blue, and the overall highest value for every metric is in blue-bold.
}
\label{tab:reducing_hallucinaion_ukabs_consistency}
\end{table}

Recall that when LegLED was used to summarize Indian or UK judgements, there were several instances of US court and statute names being hallucinated in the summaries (e.g., the examples stated in Table~\ref{tab:inconsistency-example-3} and Table~\ref{tab:inconsistency-example-4}). Importantly, we did \textit{not} observe such instances in the summaries generated by LegLED-IN or LegLED-UK. While the summaries generated by LegLED-IN or LegLED-UK also contain some issues such as words or sentences being merged together, we did not observe such gross hallucinations in the summaries generated by these fine-tuned versions.

Thus, for both datasets IN-Abs and UK-Abs, it is seen that fine-tuning over the target domain data helps to improve the quality as well as consistency of the abstractive summaries. If data in the target domain is available, it seems an effective approach to utilize such data for fine-tuning abstractive summarizers.

\subsection{Suitable prompting of LLMs}

Unlike the abstractive summarization models described previously, LLMs are very difficult/expensive to fine-tune. Hence, avoiding hallucinations in LLMs is much more challenging and is, in fact, an open question~\cite{LLM-hallucination-survey}. 

In this work, we explore a very simple approach to reducing hallucinations in LLMs -- we include in the prompt an instruction to avoid hallucinations / inconsistencies. Also, since we observed the presence of incomplete sentences in the generated summaries, we included an instruction to output complete sentences only. 
Specifically, we add the following text in the prompt -- ``\textit{Output complete sentences and not half sentences. Do not have hallucinations and inconsistencies in your summary.}''

We selected those variations of the LLMs that performed the best on the two datasets (e.g., chatgpt-explicit, davinci-summ, and chatgpt-16k-long for the UK-Abs dataset), and created new variations by adding the above-mentioned text in the prompt. For instance, we got a new variation from chatgpt-explicit, which we call \textbf{Chatgpt-reduce-hallucination} (abbreviated as \textbf{chatgpt-RH}, by using the prompt:
``Your task is to summarize the following document in at most \verb|<YY>| words. Output complete sentences and not half sentences. Do not have hallucinations and inconsistencies in your summary. The document to be summarized is given within \verb|<>|. Document to summarize - \verb|<text to summarize>|''.
Similarly, we got the variations \textbf{Davinci-reduce-hallucination} (abbreviated as \textbf{davinci-RH}) from davinci-summ, and 
\textbf{chatgpt-16k-RH-hallucination} (abbreviated as \textbf{chatgpt-16k-RH}) from chatgpt-16k-long, by including the explicit instructions for reducing hallucinations and outputing complete sentences.

\vspace{2mm}
\noindent {\bf Analysing the effect of the Semantic similarity based approach to reduce hallucination:} 
We now compare the quality of summaries generated by the SS variations of the LLMs with that of the summaries generated by the variations that achieved the best result in Section~\ref{sec:results} (e.g., chatgpt-summ with chatgpt-SS, davinci-summ with davinci-SS, gpt4-summ with gpt4-SS).

For the IN-Abs dataset, Table~\ref{tab:automated-metrics-inabs-chatgpt} shows the ROUGE, METEOR, and BERTScore metrics for the different variations of the LLMs while Table~\ref{tab:reducing_hallucinaion_inabs_consistency_general} shows the consistency metrics for the different variations.
Similarly, for the UK-Abs dataset, Table~\ref{tab:automated-metrics-ukabs-chatgpt} shows the ROUGE, METEOR, and BERTScore metrics while 
Table~\ref{tab:reducing_hallucinaion_ukabs_consistency_general} shows the consistency metrics for the different versions of the LLMs.
Across both datasets, we find that the RH versions score \textit{lower} than the `explicit' or `summ' versions of the LLMs, according to most ROUGE-based, METEOR and BertScore metrics. Whereas, the RH versions score higher according to the consistency metrics (SummaC, NEPrec, NumPrec).

\begin{table}[tb]
\centering
\begin{tabular}{lllllllll}
\hline
\textbf{Model} & \textbf{R2-P} & \textbf{R2-R} & \textbf{R2-F1} & \textbf{RL-P} & \textbf{RL-R} & \textbf{RL-F1} & \textbf{ME} & \textbf{BS}  \\ \hline

\multicolumn{9}{|c|}{\textbf{Chatgpt (Chunk size 1024)}} 
\\ \hline

chatgpt-summ &  \textcolor{blue}{0.199} & \textcolor{blue}{0.177} & \textcolor{blue}{0.186} & \textcolor{blue}{0.232} & \textbf{\textcolor{blue}{0.209}} & \textcolor{blue}{0.214} & 0.193 &  \textcolor{blue}{0.624}
\\ \hline


chatgpt-RH & 0.151
& 0.116 & 0.128 & 0.225 & 0.169 & 0.188  & \textbf{\textcolor{blue}{0.197}} & 0.540 \\ \hline

\multicolumn{9}{|c|}{\textbf{Davinci (Chunk size 1024)}} 
\\ \hline
davinci-summ & \textcolor{blue}{0.220} & \textbf{\textcolor{blue}{0.179}} & \textbf{\textcolor{blue}{0.195}} & \textcolor{blue}{0.251} & \textcolor{blue}{0.205} & \textbf{\textcolor{blue}{0.223}} & \textcolor{blue}{0.191} & 0.624 \\ \hline

davinci-RH & 0.161 & 0.146 & 0.151 & 0.205 & 0.140 & 0.167  & 0.187 & \textbf{\textcolor{blue}{0.630}} \\ \hline

\multicolumn{9}{|c|}{\textbf{Chatgpt-16k (Chunk size 8192)}} 
\\ \hline

chatgpt-16k-long & \textbf{\textcolor{blue}{0.231}} & \textcolor{blue}{0.127} & \textcolor{blue}{0.149} & \textbf{\textcolor{blue}{0.319}} & \textcolor{blue}{0.176} & \textcolor{blue}{0.206} & 0.147 & \textcolor{blue}{0.615} \\ \hline

chatgpt-16k-RH & 0.143 & 0.106 & 0.118 & 0.216 & 0.159 & 0.178  & \textcolor{blue}{0.176} & 0.568 \\ \hline

\end{tabular}

\caption{ROUGE, METEOR, BERTScore metrics for general-domain LLMs, before and after we use prompting to reduce hallucinations, over the IN-Abs dataset. The higher value for every metric is shown in blue, and the overall highest value for each metric is shown in blue-bold.
}
\label{tab:automated-metrics-inabs-chatgpt}
\end{table}

\begin{table}[tb]
\centering
\begin{tabular}{llll}
\hline
\textbf{Model} & \textbf{SummaC} & \textbf{NEPrec} & \textbf{NumPrec} \\ \hline

\multicolumn{4}{|c|}{\textbf{Chatgpt (Chunk size 1024)}} 
\\ \hline
chatgpt-summ & 0.573 & 0.901 & 0.956 \\ \hline
chatgpt-RH & \textcolor{blue}{0.600} & \textcolor{blue}{0.917} & \textcolor{blue}{0.979} \\ \hline

\multicolumn{4}{|c|}{\textbf{Davinci (Chunk size 1024)}} 
\\ \hline
davinci-summ & 0.635 & 0.895 & 0.932 \\ \hline
davinci-RH & \textcolor{blue}{0.645} & \textcolor{blue}{0.922} & \textcolor{blue}{0.964} \\ \hline

\multicolumn{4}{|c|}{\textbf{Chatgpt-16k (Chunk size 8192)}} 
\\ \hline
chatgpt-16k-long & 0.633 & 0.944 & 0.975 \\ \hline
chatgpt-16k-RH & \textbf{\textcolor{blue}{0.658}} & \textbf{\textcolor{blue}{0.955}} & \textbf{\textcolor{blue}{0.989}} \\ \hline

\end{tabular}

\caption{Consistency metrics for general-domain LLMs before and after we use prompting to reduce hallucinations, over the IN-Abs dataset. The higher value for every metric is shown in blue, and the overall highest value for every metric is in blue-bold.
}
\label{tab:reducing_hallucinaion_inabs_consistency_general}
\end{table}

\begin{table}[tb]
\centering
\begin{tabular}{lllllllll}
\hline
\textbf{Model} & \textbf{R2-P} & \textbf{R2-R} & \textbf{R2-F1} & \textbf{RL-P} & \textbf{RL-R} & \textbf{RL-F1} & \textbf{ME} & \textbf{BS}  \\ \hline

\multicolumn{9}{|c|}{\textbf{Chatgpt (Chunk size 1024)}} 
\\ \hline

chatgpt-explicit & 0.164 & \textcolor{blue}{0.163} & \textcolor{blue}{0.163} & 0.214 & \textbf{\textcolor{blue}{0.210*}} & \textcolor{blue}{0.211} & \textcolor{blue}{0.218} & \textbf{\textcolor{blue}{0.640}} \\ \hline

chatgpt-RH & \textcolor{blue}{0.169} & 0.125 & 0.140 & \textbf{\textcolor{blue}{0.245}} & 0.188 & 0.209 & 0.170 & 0.616 \\ \hline

\multicolumn{9}{|c|}{\textbf{Davinci (Chunk size 1024)}} 
\\ \hline

davinci-summ & \textbf{\textcolor{blue}{0.196}} & \textbf{\textcolor{blue}{0.181}} & \textbf{\textcolor{blue}{0.187}} & \textcolor{blue}{0.223} & \textcolor{blue}{0.206} & \textbf{\textcolor{blue}{0.213}} & \textbf{\textcolor{blue}{0.219}} & \textcolor{blue}{0.638}
\\ \hline

davinci-RH & 0.186 & 0.114 & 0.131 & 0.206 & 0.188 & 0.194 & 0.176 & 0.617 \\ \hline

\multicolumn{9}{|c|}{\textbf{Chatgpt-16k (Chunk size 8192)}} 
\\ \hline

chatgpt-16k-long & 0.153 & \textcolor{blue}{0.144} & \textcolor{blue}{0.148} & 0.213 & \textcolor{blue}{0.199} & \textcolor{blue}{0.205} & \textcolor{blue}{0.183} & \textcolor{blue}{0.619} \\ \hline

chatgpt-16k-RH & \textcolor{blue}{0.159} & 0.126 & 0.135 & \textcolor{blue}{0.215} & 0.139 & 0.160  & 0.155 & 0.588 \\ \hline

\end{tabular}

\caption{ROUGE, METEOR, BERTScore metrics for general-domain LLMs before and after we use prompting to reduce hallucinations, over the UK-Abs dataset. The higher value for every metric is shown in blue, and the overall highest value for every metric is in blue-bold.
}
\label{tab:automated-metrics-ukabs-chatgpt}
\end{table}

\begin{table}[tb]
\centering
\begin{tabular}{llll}
\hline
\textbf{Model} & \textbf{SummaC} & \textbf{NEPrec} & \textbf{NumPrec} \\ \hline

\multicolumn{4}{|c|}{\textbf{Chatgpt (Chunk size 1024)}}  \\ \hline
chatgpt-explicit & 0.622 & 0.900 & 0.952 \\ \hline
chatgpt-RH & \textcolor{blue}{0.634} & \textcolor{blue}{0.947} & \textcolor{blue}{0.981} \\ \hline

\multicolumn{4}{|c|}{\textbf{Davinci (Chunk size 1024)}}  \\ \hline
davinci-summ & 0.634 & 0.926 & 0.977 \\ \hline
davinci-RH & \textcolor{blue}{0.635} & \textcolor{blue}{0.932} & \textcolor{blue}{0.984} \\ \hline

\multicolumn{4}{|c|}{\textbf{Chatgpt-16k (Chunk size 8192)}}  \\ \hline
chatgpt-16k-long & 0.644 & 0.956 & 0.981 \\ \hline
chatgpt-16k-RH & \textbf{\textcolor{blue}{0.654}} & \textbf{\textcolor{blue}{0.967}} & \textbf{\textcolor{blue}{0.988}}  \\ \hline

\end{tabular}

\caption{Consistency metrics for general-domain LLMs before and after we use prompting to reduce hallucinations, over the UK-Abs dataset. The higher value for every metric is shown in blue, and the overall highest value for every metric is in blue-bold.
}
\label{tab:reducing_hallucinaion_ukabs_consistency_general}
\end{table}

We manually examined several summaries generated by the RH variants and the corresponding summaries by the `summ', `long' and `explicit' variants, to understand the reasons behind the fall in the summary quality of the RH variants.  
We observed that the summaries generated by the RH-variants have lesser amount of key information, such as named entities (names of persons, organizations, etc.) in the case document, as compared to those generated by the other variants. 
For instance, the average number of named entities in the summaries generated by chatgpt-RH is 13.9, while that in the summaries generated by chatgpt-summ is 16.5.
The summaries generated by the RH-variant have lesser amounts of unigram, bigram, longest common subsequence, and semantic similarity matches with the gold standard summaries as compared to the `summ' variant.
Since there is lesser amount of key information contained in the summaries generated by the RH variants, the ROUGE, METEOR, and BERTScore of these summaries are lower than those for the other variants. 



These observations imply that, while it is possible to reduce hallucinations / inconsistencies in the generated summaries by explicit prompting, there may be an associated reduction in the quality of the summary. A future direction of research would be to devise strategies for reducing hallucinations and inconsistencies in LLM-generated legal summaries, while maintaining the information quality of the summaries.

\subsection{Semantic similarity based approach for reducing hallucinations}

We now describe a novel approach that uses semantic similarity between entities to reduce hallucinations and inconsistencies in summaries.
Given a legal judgement $j$ and its summary $s$ generated by an AI model, the approach consists of the following steps:-
(i)~Consider all the named entities and numbers present in the legal judgement $j$. Let the set of named entities and numbers present in $j$ be $V_{j}$ (the set of named entities and numbers can be detected by any standard named entity recognizer; in particular, we use the popular Python Spacy library).
(ii)~Consider all the named entities and numbers present in the summary $s$ generated by the AI model. Let the set of named entities and numbers present in the summary be defined as $V_{s}$.
(iii)~Now define a set $V_{r} = V_{s} - V_{j}$. $V_{r}$ is the set of all those named entities / numbers that are present in the generated summary but are \textit{not} present in the original judgement (hallucinated entities).
(iv)~For every element in the set $V_{r}$, consider an embedding/representation from any pre-trained language model (specifically, we used the bert-base-uncased embeddings). Also, for every element in the set $V_{j}$, consider the same embedding.
(v)~Calculate the cosine similarity between the embeddings of every element in $V_{r}$ and every element in $V_{j}$. 
(vi)~Replace every element in the set $V_{r}$ with that element in $V_{j}$ which has the highest cosine similarity with this element, based on the embeddings (numbers to be replaced by numbers only). 
Basically, we replace every hallucinated named entity or number (in the generated summary) with the most semantically similar entity that is present in the input document.

The chatgpt variant which uses this semantic similarity based approach to reduce hallucinations is denoted as chatgpt-SS. Similarly, we denote the Davinci and chatgpt-16k variant which uses the semantic similarity based approach to reduce hallucinations as davinci-SS and chatgpt-16k-SS respectively.

Table~\ref{tab:inconsistency-SS-1} shows examples of two summary extracts with hallucinated named entities / numbers replaced by the correct entities using the semantic similarity based method. The first example shows the wrong monetary amount `Rs. 26,500' replaced by `Rs. 29,500'. The second example shows that a wrong judge name is replaced by the correct judge name using the semantic similarity based method.

\vspace{2mm}
\noindent 
{\bf Analysing the effect of semantic similarity to reduce hallucination:} 
We now compare the quality of summaries generated by the SS variations of the LLMs  with that of the summaries generated by the variations that achieved the best result in Section~\ref{sec:results} (e.g., chatgpt-summ with chatgpt-SS, davinci-summ with davinci-SS, chatgpt-16k-long with chatgpt-16k-SS). 
For the IN-Abs dataset, Table~\ref{tab:automated-metrics-inabs-chatgpt_ss} shows the ROUGE, METEOR, and BERTScore metrics for the different variations of the LLMs while Table~\ref{tab:reducing_hallucinaion_inabs_consistency_general_ss} shows the consistency metrics for the different variations.
Similarly, for the UK-Abs dataset, Table~\ref{tab:automated-metrics-ukabs-chatgpt_ss} shows the ROUGE, METEOR, and BERTScore metrics while 
Table~\ref{tab:reducing_hallucinaion_ukabs_consistency_general_ss} shows the consistency metrics for the different versions of the LLMs.
For both datasets, we observe an interesting trend. 
The SS variations not only achieve higher scores for the consistency metrics (e.g., chatgpt-SS has higher SummaC, NEPrec and NumPrec scores than chatgpt-summ in Table~\ref{tab:reducing_hallucinaion_inabs_consistency_general_ss}), but also 
mostly score higher in terms of the ROUGE, METEOR and BERTScore metrics, as compared to the other variations (e.g., chatgpt-SS has higher scores than chatgpt-summ as per most metrics in Table~\ref{tab:automated-metrics-inabs-chatgpt_ss}). 



\begin{table*}[tb]
\centering
\small
\begin{tabular}{|p{0.01\textwidth}|p{0.1\textwidth}|p{0.38\textwidth}|p{0.38\textwidth}|}
\hline
\textbf{id} & \textbf{Model} & \textbf{Extract from summary showing hallucinated named entity / number} & \textbf{Extract from corrected summary}  \\ \hline
1
&
davinci-summ
& 
The prosecution asserts that the accused initially demanded Rs. 30,000 later reduced to Rs. \textcolor{red}{26,500} for granting the complainant vacant possession of a flat. 
&
The prosecution asserts that the accused initially demanded Rs. 30,000 later reduced to Rs. \textcolor{red}{29,500} for granting the complainant vacant possession of a flat. 

\\ \hline





2
&
chatgpt-summ
& 
On February 1, 2020, the Honorable \textcolor{red}{Chandrasekhar A. Lama} of India entered a final judgment .......
&
On February 1, 2020, the Honorable \textcolor{red}{Aiyar, N. Chandrasekhara} of India entered a final judgment .......

\\ \hline

\end{tabular}
\caption{
Examples of errors in abstractive summaries generated by different summarization models for the Indian Court judgement available at \url{https://indiankanoon.org/doc/1801104/}. The third column shows the summary extract with the hallucinated named entity / number in red-colored font. The fourth column shows the same summary extract corrected by the semantic similarity based method, with the corrected named entity / number in red. 
}
\label{tab:inconsistency-SS-1}
\end{table*}

\begin{table}[tb]
\centering
\begin{tabular}{lllllllll}
\hline
\textbf{Model} & \textbf{R2-P} & \textbf{R2-R} & \textbf{R2-F1} & \textbf{RL-P} & \textbf{RL-R} & \textbf{RL-F1} & \textbf{ME} & \textbf{BS}  \\ \hline

\multicolumn{9}{|c|}{\textbf{Chatgpt (Chunk size 1024)}} 
\\ \hline

chatgpt-summ &  0.199 & 0.177 & 0.186 & 0.232 & 0.209 & 0.214 & 0.193 &  0.624
\\ \hline


chatgpt-SS & \textcolor{blue}{0.206}
& \textcolor{blue}{0.184} & \textcolor{blue}{0.192} & \textcolor{blue}{0.239} & \textbf{\textcolor{blue}{0.227}} & \textcolor{blue}{0.230}  & \textbf{\textcolor{blue}{0.197}} & \textbf{\textcolor{blue}{0.636}} \\ \hline

\multicolumn{9}{|c|}{\textbf{Davinci (Chunk size 1024)}} 
\\ \hline
davinci-summ & \textcolor{blue}{0.220} & 0.179 & 0.195 & \textcolor{blue}{0.251} & 0.205 & 0.223 & 0.191 & 0.624 \\ \hline

davinci-SS & 0.216 & \textbf{\textcolor{blue}{0.185}} & \textbf{\textcolor{blue}{0.198}} & 0.242 & \textcolor{blue}{0.213} & \textcolor{blue}{0.225} & \textcolor{blue}{0.196} & \textcolor{blue}{0.633} \\ \hline

\multicolumn{9}{|c|}{\textbf{Chatgpt-16k (Chunk size 8192)}} 
\\ \hline

chatgpt-16k-long & 0.231 & \textcolor{blue}{0.137} & 0.149 & 0.319 & 0.176 & 0.206 & \textcolor{blue}{0.154} & 0.615 \\ \hline

chatgpt-16k-SS & \textbf{\textcolor{blue}{0.242}} & 0.132 & \textcolor{blue}{0.155} & \textbf{\textcolor{blue}{0.329}} & \textcolor{blue}{0.182} & \textbf{\textcolor{blue}{0.234}} & 0.147 & \textcolor{blue}{0.624} \\ \hline

\end{tabular}

\caption{ROUGE, METEOR, BERTScore metrics for general-domain LLMs, on using semantic similarity-based approach to reduce hallucinations, over the IN-Abs dataset. The higher value for every metric is shown in blue, and the overall highest value for each metric is shown in blue-bold.
}
\label{tab:automated-metrics-inabs-chatgpt_ss}
\end{table}

\begin{table}[tb]
\centering
\begin{tabular}{llll}
\hline
\textbf{Model} & \textbf{SummaC} & \textbf{NEPrec} & \textbf{NumPrec} \\ \hline

\multicolumn{4}{|c|}{\textbf{Chatgpt (Chunk size 1024)}} 
\\ \hline
chatgpt-summ & 0.573 & 0.901 & 0.956 \\ \hline
chatgpt-SS & \textcolor{blue}{0.605} & \textcolor{blue}{0.929} & \textcolor{blue}{0.984} \\ \hline

\multicolumn{4}{|c|}{\textbf{Davinci (Chunk size 1024)}} 
\\ \hline
davinci-summ & 0.635 & 0.895 & 0.932 \\ \hline
davinci-SS & \textcolor{blue}{0.647} & \textcolor{blue}{0.932} & \textcolor{blue}{0.972} \\ \hline

\multicolumn{4}{|c|}{\textbf{Chatgpt-16k (Chunk size 8192)}} 
\\ \hline
chatgpt-16k-long & 0.633 & 0.944 & 0.975 \\ \hline
chatgpt-16k-SS & \textbf{\textcolor{blue}{0.659}} & \textbf{\textcolor{blue}{0.958}} & \textbf{\textcolor{blue}{0.993}} \\ \hline

\end{tabular}

\caption{Consistency metrics for general-domain LLMs on using semantic similarity based approach to reduce hallucinations, over the IN-Abs dataset. The higher value for every metric is shown in blue, and the overall highest value for every metric is in blue-bold.
}
\label{tab:reducing_hallucinaion_inabs_consistency_general_ss}
\end{table}

\begin{table}[tb]
\centering
\begin{tabular}{lllllllll}
\hline
\textbf{Model} & \textbf{R2-P} & \textbf{R2-R} & \textbf{R2-F1} & \textbf{RL-P} & \textbf{RL-R} & \textbf{RL-F1} & \textbf{ME} & \textbf{BS}  \\ \hline

\multicolumn{9}{|c|}{\textbf{Chatgpt (Chunk size 1024)}} 
\\ \hline

chatgpt-explicit & 0.164 & 0.163 & 0.163 & 0.214 & 0.210* & 0.211 & 0.218 & 0.640 \\ \hline

chatgpt-SS & \textcolor{blue}{0.166} & \textcolor{blue}{0.173} & \textcolor{blue}{0.167} & \textcolor{blue}{0.223} & \textbf{\textcolor{blue}{0.215*}} & \textcolor{blue}{0.218} & \textcolor{blue}{0.228} & \textbf{\textcolor{blue}{0.653}} \\ \hline

\multicolumn{9}{|c|}{\textbf{Davinci (Chunk size 1024)}} 
\\ \hline

davinci-summ & 0.196 & 0.181 & 0.187 & 0.223 & 0.206 & 0.213 & 0.219 & \textcolor{blue}{0.643}
\\ \hline

davinci-SS & \textbf{\textcolor{blue}{0.209}} & \textbf{\textcolor{blue}{0.194}} & \textbf{\textcolor{blue}{0.192}} & \textbf{\textcolor{blue}{0.236}} & \textcolor{blue}{0.212} & \textbf{\textcolor{blue}{0.220}} & \textbf{\textcolor{blue}{0.231}} & 0.638 \\ \hline

\multicolumn{9}{|c|}{\textbf{Chatgpt-16k (Chunk size 8192)}} 
\\ \hline

chatgpt-16k-long & 0.153 & \textcolor{blue}{0.146} & 0.148 & 0.213 & 0.199 & 0.205 & 0.183 & 0.619 \\ \hline

chatgpt-16k-SS & \textcolor{blue}{0.164} & 0.145 & \textcolor{blue}{0.158} & \textcolor{blue}{0.226} & \textcolor{blue}{0.201} & \textcolor{blue}{0.209}  & \textcolor{blue}{0.194} & \textcolor{blue}{0.625} \\ \hline

\end{tabular}

\caption{ROUGE, METEOR, BERTScore metrics for general-domain LLMs on using semantic similarity based approach to reduce hallucinations, over the UK-Abs dataset. The higher value for every metric is shown in blue, and the overall highest value for every metric is in blue-bold.
}
\label{tab:automated-metrics-ukabs-chatgpt_ss}
\end{table}

\begin{table}[tb]
\centering
\begin{tabular}{llll}
\hline
\textbf{Model} & \textbf{SummaC} & \textbf{NEPrec} & \textbf{NumPrec} \\ \hline

\multicolumn{4}{|c|}{\textbf{Chatgpt (Chunk size 1024)}}  \\ \hline
chatgpt-explicit & 0.622 & 0.900 & 0.952 \\ \hline
chatgpt-SS & \textcolor{blue}{0.635} & \textcolor{blue}{0.957} & \textcolor{blue}{0.986} \\ \hline

\multicolumn{4}{|c|}{\textbf{Davinci (Chunk size 1024)}}  \\ \hline
davinci-summ & 0.634 & 0.926 & 0.977 \\ \hline
davinci-SS & \textcolor{blue}{0.636} & \textcolor{blue}{0.937} & \textcolor{blue}{0.988} \\ \hline

\multicolumn{4}{|c|}{\textbf{Chatgpt-16k (Chunk size 8192)}}  \\ \hline
chatgpt-16k-long & 0.644 & 0.956 & 0.981 \\ \hline
chatgpt-16k-SS & \textbf{\textcolor{blue}{0.658}} & \textbf{\textcolor{blue}{0.969}} & \textbf{\textcolor{blue}{0.992}}  \\ \hline

\end{tabular}

\caption{Consistency metrics for general-domain LLMs on using semantic similarity based approach to reduce hallucinations, over the UK-Abs dataset. The higher value for every metric is shown in blue, and the overall highest value for every metric is in blue-bold.
}
\label{tab:reducing_hallucinaion_ukabs_consistency_general_ss}
\end{table}

These observations imply that, there is a reduction in hallucinations / inconsistencies in the generated summaries by this approach, and there is also an improvement in terms of results of the automated metrics.
Hence, the semantic similarity-based approach is quite effective in improving the quality of generated summaries. 
However, we have noticed that all hallucinations / inconsistencies cannot be corrected by this approach; hence the correction of hallucination/inconsistency in abstractive summaries remains an open area of research.

\section{Summarization of Other Types of Legal Documents}

Till now, we have applied all summarization models over two datasets of legal case judgements. In this section, we check how well the models perform over other types of legal documents.  For this, we use the GOVREPORT dataset~\cite{huang-etal-2021-efficient}.

\subsection{GOVREPORT Dataset}

The GOVREPORT dataset, obtained from~\cite{huang-etal-2021-efficient}, comprises of 19,466 detailed reports from the U.S. Government Accountability Office (GAO) and the Congressional Research Service (CRS). These reports, prepared in response to congressional requests, span a broad range of national policy topics and include human-written summaries. 
During data collection, some boiler-plate parts were removed from the crawled files, and the section and paragraph structure of the documents and summaries were retained. 
The dataset features 12,228 GAO reports and 7,238 CRS reports.
On average, GAO reports have 6.9 sections, while CRS reports have 4.6 sections. The dataset is organized 
 into training, validation, and test sets based on the publication dates, resulting in 17,519 training samples, 974 validation samples, and 973 test samples.
Table~\ref{tab:govreport-stats} shows the dataset statistics for the GOVREPORT dataset.
Following \cite{grusky-etal-2018-newsroom,shen2022multi}, the coverage and density of the legal judgements with respect to the gold-standard summaries is 0.16 and 1.40 respectively for this dataset.

In this work, the summarization models are evaluated over a randomly sampled 100 document-summary pairs from the test set of GOVREPORT. The supervised summarization models are trained over the training set of GOVREPORT.

\subsection{Summarization results on GOVREPORT dataset}

We now study the performances of summarization models on the GOVREPORT dataset. To this end, we report the results of some of the best performing LLMs and summarization models (as observed in the previous sections of this paper).
From the GPT-4 Turbo variants, we report the results of the best performing variant gpt4-summ. 
Similarly, from the Llama2-70b and Chatgpt variants, we report the results of llama-summ and chatgpt-summ.
We also report the results of LegPegasus and LegLED fine-tuned on the training dataset of GOVREPORT; we denote these models as LegPegasus-GR and LegLED-GR. 
We also report the results of best performing extractive summarization model namely CaseSummarizer.


\begin{table}[tb]
    \centering
\begin{tabular}{|p{0.1\columnwidth}|p{0.15\columnwidth}|p{0.25\columnwidth}|p{0.3\columnwidth}|}
\hline
 & \textbf{Nos. of documents} & \textbf{Average nos. of words per document} & \textbf{Average nos. of words per gold-standard summary} 
\\ \hline 
Train set & 17519 & 9407.17 &  556.75 \\ \hline
Test set & 973 & 9409.45 & 540.48 \\ \hline
Validation set & 974 & 9417.19 & 545.04 \\ \hline

\end{tabular}
\caption{\textbf{Dataset statistics for GOVREPORT dataset.}
To get the average number of words per document (or summary), we first calculate the number of words in every document (summary). Then we add up the number of words from all documents (summaries), and divide the total word-count by the number of documents (summaries).}

\label{tab:govreport-stats}
\end{table}

\begin{table*}[tb]
\small
\centering
\scalebox{0.9}
{
\begin{tabular}{l|lll|lll|ll}
\hline
\textbf{Model} & \textbf{R2-P} & \textbf{R2-R} & \textbf{R2-F1} & \textbf{RL-P} & \textbf{RL-R} & \textbf{RL-F1} & \textbf{ME} & \textbf{BS}  
\\
\hline

\multicolumn{9}{|c|}{\textbf{General domain LLMs}} \\ \hline

gpt4-summ & 0.386 & 0.169 & 0.205 & 0.453 & 0.292 & 0.326 & 0.205 & 0.627 \\ \hline

chatgpt-summ & \textcolor{blue}{0.395} & \textbf{\textcolor{blue}{0.172}} & \textbf{\textcolor{blue}{0.221}} & \textcolor{blue}{0.461} & \textbf{\textcolor{blue}{0.302}} & \textbf{\textcolor{blue}{0.346}} & \textcolor{blue}{0.215}  & \textbf{\textcolor{blue}{0.629}} \\ \hline

llama-summ & 0.369 & 0.148 & 0.179 & 0.433 & 0.283 & 0.317 & 0.202 & 0.602 \\ \hline













\multicolumn{9}{|c|}{\textbf{Legal domain-specific abstractive models}} \\ \hline

LegPegasus-GR & \textbf{\textcolor{blue}{0.408}} & \textcolor{blue}{0.149} & \textcolor{blue}{0.193} & \textbf{\textcolor{blue}{0.481}} & \textcolor{blue}{0.215} & \textcolor{blue}{0.304} & \textbf{\textcolor{blue}{0.217}}  & \textcolor{blue}{0.618} \\ \hline

LegLED-GR & 0.382 & 0.147 & 0.184 & 0.460 & 0.204 & 0.298 & 0.205  & 0.608 \\ \hline






\multicolumn{9}{|c|}{\textbf{Extractive models}} 
\\ \hline

CaseSummarizer & 0.351 & 0.126 & 0.183 & 0.431 & 0.188 & 0.296 & 0.201  & 0.604\\ \hline

\end{tabular}
}
\caption{ROUGE, METEOR, BERTScore metrics for best performing general-domain LLMs, abstractive and extractive summarization models on the GOVREPORT dataset.
The value shown in blue is the best value in every summarization family for every metric. The value shown in blue-bold is the highest value for every metric.
}
\label{tab:in-abs_extractive_abstrative-first-govreport}
\end{table*}

\begin{table}[tb]
\centering
\begin{tabular}{llll}
\hline
\textbf{Model} & \textbf{SummaC} & \textbf{NEPrec} & \textbf{NumPrec} \\ \hline

\multicolumn{4}{|c|}{\textbf{General domain LLMs}} \\ \hline

gpt4-summ & \textcolor{blue}{0.599} & \textbf{\textcolor{blue}{0.952}} & \textcolor{blue}{0.984}  \\ \hline





chatgpt-summ & 0.579 & 0.927 & 0.946  \\ \hline

llama-summ & 0.564 & 0.872 & 0.873  \\ \hline













\multicolumn{4}{|c|}{\textbf{Legal Domain-specific abstractive models}} 
\\ \hline

LegPegasus-GR & \textbf{\textcolor{blue}{0.617}} & \textcolor{blue}{0.935} & \textbf{\textcolor{blue}{0.986}} \\ \hline

LegLED-GR & 0.603 & 0.904 & 0.973 \\ \hline

\end{tabular}

\caption{Consistency metrics for different abstractive summarization models on GOVREPORT dataset. The value shown in blue is the best value in every summarization family for every metric. The value shown in bold is the highest value for every metric.
}
\label{tab:consistency-metrics-inabs-prompting-govreport}
\end{table}

\vspace{2mm}
\noindent {\bf Comparing general-domain LLMs with domain-specific abstractive summarizers and extractive models:}
Table~\ref{tab:in-abs_extractive_abstrative-first-govreport} shows the ROUGE, METEOR, BERTScore metrics for the aforementioned summarization models on the GOVREPORT dataset.
Among the legal domain-specific abstractive models, LegPegasus-GR performs the best.
Among the General domain LLMs, chatgpt-summ performs the best for most automated metrics.
We see that both LLMs and legal domain-specific abstractive models perform much better than the extractive CaseSummarizer over this dataset. This is somewhat expected, since CaseSummarizer is specifically meant for summarizing legal case judgements and not other types of legal documents.

In general, we see higher metric values, especially for the ROUGE scores, for GOVREPORT than what we obtained for the IN-Abs and UK-Abs datasets. These trends seem to indicate that summarizing legal case judgement reports is a more complicated task than summarizing the Government reports in the GOVREPORT dataset.

\vspace{2mm}
\noindent {\bf Consistency of generated summaries:}
Table~\ref{tab:consistency-metrics-inabs-prompting-govreport} shows the consistency metrics for summaries generated by different abstractive models on the GOVREPORT dataset.
The value shown in blue is the best value in every summarization family for every metric. The value shown in blue-bold is the overall highest value for every metric.
gpt4-summ gets the highest NEPrec score, while LegPegasus-GR achieves the highest SummaC and NumPrec scores.


\section{Human evaluation of summaries} \label{sec:human-eval}

In this final section, we perform a human evaluation of the summaries generated by some of the best-performing models. 
For this, we consulted three senior Law students from the Rajiv Gandhi School on Intellectual Property Law, a reputed Law school in India. Due to their limited availability, we considered only the five best-performing summarization models for each dataset. Specifically, we considered the best-performing model from every summarization family. 

For the IN-Abs dataset, we considered chatgpt-summ (the best-performing model amongst Chatgpt-1024 models), davinci-summ (the best-performing model amongst Davinci-1024 models), chatgpt-16k-long (the best-performing model amongst Chatgpt-16k models), CaseSummarizer (the best-performing extractive model), and LegLED-IN (the best-performing model amongst legal domain-specific abstractive summarization models).
For the UK-Abs dataset, we considered chatgpt-explicit (the best-performing model amongst Chatgpt-1024 models), davinci-summ (the best-performing model amongst Davinci-1024 models), chatgpt-16k-long (the best-performing model amongst Chatgpt-16k models), CaseSummarizer (the best-performing extractive model), and LegPegasus-UK (the best-performing model amongst legal domain-specific abstractive models).
For the GOVREPORT dataset, we considered for human evaluation chatgpt-summ, llama-summ, gpt4-summ,  LegPegasus-GR, and CaseSummarizer.

\vspace{2mm}
\noindent {\bf Metrics for human evaluation:} We used the following four metrics for the human evaluation of the summaries:-

\noindent $\bullet$ \textbf{Informativeness:}
This metric measures how much relevant information the summary contains from the source document. The annotators were asked to assess if the summary captures the essential points of the original document. A higher value of informativeness means a higher-quality summary.

\noindent $\bullet$ \textbf{Redundancy:}
This metric measures how much information is repeated in the summary. Redundancy can negatively impact the quality of a summary. A {\it lower} value of redundancy means a higher-quality summary.

\noindent $\bullet$ \textbf{Factuality:}
This metric determines if the information presented in the summary is factually accurate. The annotators were asked to compare the summary content with the source document to assess accuracy. A higher value of factuality means a higher-quality summary.

\noindent $\bullet$ \textbf{Coherence:}
This metric assesses if the sentences in the summary are logically connected and form a coherent and understandable narrative. A higher value of Coherence means a higher-quality summary.

\vspace{2mm}
\noindent {\bf Method of human evaluation:}
From each dataset, we considered 25 randomly selected documents, and their summaries generated by the 5 models stated above. 
We asked the human annotators to give a score between 1 and 5 to every model-generated summary, for each of the above metrics. 
Each annotator was asked to evaluate the summaries independently, i.e., without discussing with the others. Also, the annotators were {\it not} told which summary was generated by which model, in order to ensure an unbiased evaluation.
Through this evaluation, a particular summarization model got 75 scores (scores provided to model-generated summaries for 25 documents, by 3 annotators) for every metric, and then we take the mean/average of the 75 scores. We report these average scores for each metric, separately for the three datasets.

\begin{table}[tb]
\centering
\begin{tabular}{l|cccc}
\hline
\textbf{Model} & \textbf{Informativeness} & \textbf{Redundancy} & \textbf{Factuality} & \textbf{Coherence} \\ \hline

chatgpt-explicit  &  3.34 & 1.65 & 3.68 & 3.26 \\ \hline

davinci-summ  & \textcolor{blue}{\textbf{3.67}} & 1.53 & 3.73 & 3.22 \\ \hline

chatgpt-16k-long &  3.01 & \textcolor{blue}{\textbf{1.28}} & \textcolor{blue}{\textbf{4.06}} & \textcolor{blue}{\textbf{3.67}}  \\ \hline

CaseSummarizer & 3.09 & 1.54 & 3.62 & 3.13 \\ \hline

LegPegasus-UK & 3.53 & 1.61 & 3.90 & 3.44 \\ \hline

\end{tabular}

\caption{Human evaluation scores on UK-Abs dataset calculated by averaging the scores provided by three annotators for 25 summaries for every model. The highest scores for Informativeness, Factuality, and Coherence, and the lowest score for Redundancy are shown in blue-bold.}
\label{tab:human_evaluation_ukabs}
\end{table}

\begin{table}[tb]
\centering
\begin{tabular}{l|cccc}
\hline
\textbf{Model} & \textbf{Informativeness} & \textbf{Redundancy} & \textbf{Factuality} & \textbf{Coherence} \\ \hline


chatgpt-summ  &  3.34 & 1.66 & 3.46 & 3.26 \\ \hline

davinci-summ  & 3.46 & 1.60 & 3.49 & 3.48 \\ \hline

chatgpt-16k-long &  2.98 & \textcolor{blue}{\textbf{1.30}} & \textcolor{blue}{\textbf{4.20}} & \textcolor{blue}{\textbf{3.89}}  \\ \hline

CaseSummarizer & 3.54 & 1.74 & 3.64 & 3.52 \\ \hline

LegLED-IN & \textcolor{blue}{\textbf{3.61}} & 1.68 & 3.85 & 3.61 \\ \hline

\end{tabular}

\caption{Human evaluation scores on IN-Abs dataset calculated by averaging the scores provided by three annotators for 25 summaries for every model. The highest scores for Informativeness, Factuality, and Coherence, and the lowest score for Redundancy are shown in blue-bold.}
\label{tab:human_evaluation_inabs}
\end{table}

\begin{table}[tb]
\centering
\begin{tabular}{l|cccc}
\hline
\textbf{Model} & \textbf{Informativeness} & \textbf{Redundancy} & \textbf{Factuality} & \textbf{Coherence} \\ \hline


chatgpt-summ  &  3.25 & 1.60 & 3.43 & 3.26 \\ \hline

llama-summ  & 3.16 & 1.56 & 3.40 & 3.04 \\ \hline

gpt4-summ &  2.94 & \textcolor{blue}{\textbf{1.35}} & \textcolor{blue}{\textbf{3.89}} & \textcolor{blue}{\textbf{3.94}}  \\ \hline

CaseSummarizer &  3.10 & 1.61 & 3.26 & 3.10 \\ \hline

LegPegasus-GR & \textcolor{blue}{\textbf{3.56}} & 1.62 & 3.56 & 3.47 \\ \hline

\end{tabular}

\caption{Human evaluation scores on GOVREPORT dataset calculated by averaging the scores provided by three annotators for 25 summaries for every model. The highest scores for Informativeness, Factuality, and Coherence, and the lowest score for Redundancy are shown in blue-bold.}
\label{tab:human_evaluation_govreport}
\end{table}

\vspace{2mm}
\noindent {\bf Inter-annotator agreement:}
We measure the inter-annotator agreement scores in terms of Fleiss Kappa over the scores given by the human annotators, over all the datasets combined.
The Fleiss Kappa for Informativeness, Redundancy, Factuality and Coherence come out to be 0.54, 0.65, 0.42, and 0.56 respectively.
The Fleiss Kappa scores for Informativeness, Factuality, and Coherence represent moderate agreement while the Fleiss Kappa score for Redundancy represents substantial agreement~\cite{landis1977measurement}.

\vspace{2mm}
\noindent {\bf Results of human evaluation:}
Table~\ref{tab:human_evaluation_ukabs} shows the human evaluation scores for the UK-Abs dataset, Table~\ref{tab:human_evaluation_inabs} for the IN-Abs dataset, and
Table~\ref{tab:human_evaluation_govreport} shows the human evaluation scores for the GOVREPORT dataset.
For IN-Abs and UK-Abs datasets, the summaries generated by chatgpt-16k-long have been given the best scores by the annotators, in terms of Redundancy, Factuality and Coherence. 
In terms of Informativeness, davinci-summ has the highest score for UK-Abs while LegLED-IN gets the highest score for IN-Abs. 
For GOVREPORT dataset, gpt4-summ has the best scores in terms of  redundancy, factuality, and coherence, while LegPegasus-GR has the highest informativeness.


These scores reflect the trade-off in breaking up a long legal document into chunks and independently summarizing every chunk (which is the approach for every other summarization model, other than chatgpt-16k-long, and gpt4-summ). 
The chunking approach potentially leads to redundancy (e.g., if the same information comes into the summary from multiple chunks) and lack of coherence at the chunk-boundaries. 
In contrast, chatgpt-16k-long (with a chunk size of 8,192) and GPT4 can accommodate most case judgements and government reports as a whole, and hence can generate summaries with lesser redundancy and higher coherence.
On the other hand, informativeness of the summaries are better with the chunking strategy, since more information from each chunk can be accommodated in the summaries. This is possibly why davinci-summ, LegLED-IN and LegPegasus-GR are given the highest scores for Informativeness by the Law students.

\section{Concluding Discussion}
\label{sec:conclu}

To our knowledge, this is the first work that systematically compares the performances of three different families of models for the practical and challenging task of legal case judgement summarization --
(1)~traditional extractive summarization models, (2)~domain-specific abstractive summarization models (Legal-Pegasus and Legal-LED), and (3)~general domain LLMs such as ChatGPT, Davinci, Llama2-70b and GPT-4-Turbo. 
We conduct comprehensive experiments over three datasets from the Indian, UK Supreme Courts and GOVREPORT dataset. 
Our experiments lead to the following insights. 

\begin{itemize}
    
    
    \item Abstractive models and LLMs generally outperform extractive models in terms of both quantitative metrics as well as human evaluation. In particular, LLMs like Text-Davinci-003, Turbo-GPT-3.5, Llama-70b and GPT4 perform well even without specific legal document training. However, generative models often contain hallucinations and inconsistencies in the generated summaries.

    \item Fine-tuning abstractive models with the target domain data, if available, helps in reducing hallucinations/inconsistencies as well as improves the quality of the summaries (as seen on both UK-Abs and IN-Abs datasets).

    \item Suitable prompting of LLMs can help in reducing hallucinations/inconsistencies in the generated summaries. But there may be an associated reduction of the summary quality. An important future direction of research would be to reduce inconsistencies in LLM-generated summaries while maintaining the summary quality. 

    \item We discussed a semantic similarity-based approach to reduce hallucinations in the summaries generated by the LLMs. However, complex errors, such as confusion between names or numbers are difficult to detect or prevent completely.

    \item The extreme length of legal case judgements is a domain-specific challenge. There is a trade-off associated with the approach of breaking these long documents into chunks. As seen from the human evaluation, and as already demonstrated in \cite{moro2022semantic}, chunking leads to better information quality in the summaries, but also leads to redundancy and lack of coherence. 
\end{itemize}

Based on these results, we conclude that, for complex domains like law, LLMs and pre-trained abstractive summarization models are not ready yet for fully automatic deployment. 
A human-in-the-loop approach, where a legal expert monitors the generated summaries, may be more appropriate. Furthermore, better methods are needed to detect complex errors in abstractive summaries. We plan to
explore these directions in the future.

The present work has certain limitations. We have used a basic segmentation technique to manage lengthy documents, which has certain limitations. First, the token-level segmentation strategy can cause the last sentence of a chunk to end abruptly, potentially affecting readability. Second, within a single chunk, sentences might address different topics, which can reduce the coherence of the chunks. 
While this chunking aspect is not the main focus of the present study, it is important to acknowledge that the quality of the chunks significantly influences the final findings and conclusions of the work. An important future direction of research is to explore better segmentation strategies for handling long legal documents.

\section*{Acknowledgements}

The authors acknowledge the anonymous reviewers whose comments greatly helped to improve the paper, and the Law students who helped with the human evaluation of the summaries.
The work is partially supported by the Technology Innovation Hub (TIH) on AI for Interdisciplinary Cyber-Physical Systems (AI4ICPS) set up by IIT Kharagpur under the aegis of DST, Government of India (GoI).


\bibliography{sn-bibliography}


\begin{thebibliography}{50}
\ifx \bisbn   \undefined \def \bisbn  #1{ISBN #1}\fi
\ifx \binits  \undefined \def \binits#1{#1}\fi
\ifx \bauthor  \undefined \def \bauthor#1{#1}\fi
\ifx \batitle  \undefined \def \batitle#1{#1}\fi
\ifx \bjtitle  \undefined \def \bjtitle#1{#1}\fi
\ifx \bvolume  \undefined \def \bvolume#1{\textbf{#1}}\fi
\ifx \byear  \undefined \def \byear#1{#1}\fi
\ifx \bissue  \undefined \def \bissue#1{#1}\fi
\ifx \bfpage  \undefined \def \bfpage#1{#1}\fi
\ifx \blpage  \undefined \def \blpage #1{#1}\fi
\ifx \burl  \undefined \def \burl#1{\textsf{#1}}\fi
\ifx \doiurl  \undefined \def \doiurl#1{\url{https://doi.org/#1}}\fi
\ifx \betal  \undefined \def \betal{\textit{et al.}}\fi
\ifx \binstitute  \undefined \def \binstitute#1{#1}\fi
\ifx \binstitutionaled  \undefined \def \binstitutionaled#1{#1}\fi
\ifx \bctitle  \undefined \def \bctitle#1{#1}\fi
\ifx \beditor  \undefined \def \beditor#1{#1}\fi
\ifx \bpublisher  \undefined \def \bpublisher#1{#1}\fi
\ifx \bbtitle  \undefined \def \bbtitle#1{#1}\fi
\ifx \bedition  \undefined \def \bedition#1{#1}\fi
\ifx \bseriesno  \undefined \def \bseriesno#1{#1}\fi
\ifx \blocation  \undefined \def \blocation#1{#1}\fi
\ifx \bsertitle  \undefined \def \bsertitle#1{#1}\fi
\ifx \bsnm \undefined \def \bsnm#1{#1}\fi
\ifx \bsuffix \undefined \def \bsuffix#1{#1}\fi
\ifx \bparticle \undefined \def \bparticle#1{#1}\fi
\ifx \barticle \undefined \def \barticle#1{#1}\fi
\bibcommenthead
\ifx \bconfdate \undefined \def \bconfdate #1{#1}\fi
\ifx \botherref \undefined \def \botherref #1{#1}\fi
\ifx \url \undefined \def \url#1{\textsf{#1}}\fi
\ifx \bchapter \undefined \def \bchapter#1{#1}\fi
\ifx \bbook \undefined \def \bbook#1{#1}\fi
\ifx \bcomment \undefined \def \bcomment#1{#1}\fi
\ifx \oauthor \undefined \def \oauthor#1{#1}\fi
\ifx \citeauthoryear \undefined \def \citeauthoryear#1{#1}\fi
\ifx \endbibitem  \undefined \def \endbibitem {}\fi
\ifx \bconflocation  \undefined \def \bconflocation#1{#1}\fi
\ifx \arxivurl  \undefined \def \arxivurl#1{\textsf{#1}}\fi
\csname PreBibitemsHook\endcsname

\bibitem[\protect\citeauthoryear{Deroy et~al.}{2023}]{deroy2023ensemble}
\begin{botherref}
\oauthor{\bsnm{Deroy}, \binits{A.}},
\oauthor{\bsnm{Ghosh}, \binits{K.}},
\oauthor{\bsnm{Ghosh}, \binits{S.}}:
Ensemble methods for improving extractive summarization of legal case judgements.
Artificial Intelligence and Law,
1--59
(2023)
\end{botherref}
\endbibitem

\bibitem[\protect\citeauthoryear{Bhattacharya et~al.}{2021}]{bhattacharya2021incorporating}
\begin{bchapter}
\bauthor{\bsnm{Bhattacharya}, \binits{P.}},
\bauthor{\bsnm{Poddar}, \binits{S.}},
\bauthor{\bsnm{Rudra}, \binits{K.}},
\bauthor{\bsnm{Ghosh}, \binits{K.}},
\bauthor{\bsnm{Ghosh}, \binits{S.}}:
\bctitle{Incorporating domain knowledge for extractive summarization of legal case documents}.
In: \bbtitle{Proc. International Conference on Artificial Intelligence and Law (ICAIL)},
pp. \bfpage{22}--\blpage{31}
(\byear{2021})
\end{bchapter}
\endbibitem

\bibitem[\protect\citeauthoryear{Polsley et~al.}{2016}]{polsley-etal-2016-casesummarizer}
\begin{bchapter}
\bauthor{\bsnm{Polsley}, \binits{S.}},
\bauthor{\bsnm{Jhunjhunwala}, \binits{P.}},
\bauthor{\bsnm{Huang}, \binits{R.}}:
\bctitle{{C}ase{S}ummarizer: A system for automated summarization of legal texts}.
In: \bbtitle{Proceedings of {COLING} 2016, the 26th International Conference on Computational Linguistics: System Demonstrations},
pp. \bfpage{258}--\blpage{262}
(\byear{2016})
\end{bchapter}
\endbibitem

\bibitem[\protect\citeauthoryear{Zhong et~al.}{2019}]{10.1145/3322640.3326728}
\begin{bchapter}
\bauthor{\bsnm{Zhong}, \binits{L.}},
\bauthor{\bsnm{Zhong}, \binits{Z.}},
\bauthor{\bsnm{Zhao}, \binits{Z.}},
\bauthor{\bsnm{Wang}, \binits{S.}},
\bauthor{\bsnm{Ashley}, \binits{K.D.}},
\bauthor{\bsnm{Grabmair}, \binits{M.}}:
\bctitle{Automatic summarization of legal decisions using iterative masking of predictive sentences}.
In: \bbtitle{Proceedings of the Seventeenth International Conference on Artificial Intelligence and Law (ICAIL)},
pp. \bfpage{163}--\blpage{172}
(\byear{2019})
\end{bchapter}
\endbibitem

\bibitem[\protect\citeauthoryear{Deroy et~al.}{2024}]{deroy2024artificialintelligenceailegal}
\begin{botherref}
\oauthor{\bsnm{Deroy}, \binits{A.}},
\oauthor{\bsnm{Bailung}, \binits{N.K.}},
\oauthor{\bsnm{Ghosh}, \binits{K.}},
\oauthor{\bsnm{Ghosh}, \binits{S.}},
\oauthor{\bsnm{Chakraborty}, \binits{A.}}:
Artificial Intelligence (AI) in Legal Data Mining
(2024).
\url{https://arxiv.org/abs/2405.14707}
\end{botherref}
\endbibitem

\bibitem[\protect\citeauthoryear{Shukla et~al.}{2022}]{shukla2022legal}
\begin{bchapter}
\bauthor{\bsnm{Shukla}, \binits{A.}},
\bauthor{\bsnm{Bhattacharya}, \binits{P.}},
\bauthor{\bsnm{Poddar}, \binits{S.}},
\bauthor{\bsnm{Mukherjee}, \binits{R.}},
\bauthor{\bsnm{Ghosh}, \binits{K.}},
\bauthor{\bsnm{Goyal}, \binits{P.}},
\bauthor{\bsnm{Ghosh}, \binits{S.}}:
\bctitle{Legal case document summarization: Extractive and abstractive methods and their evaluation}.
In: \bbtitle{Proceedings of the Conference of the Asia-Pacific Chapter of the Association for Computational Linguistics and the International Joint Conference on Natural Language Processing (Volume 1: Long Papers)},
pp. \bfpage{1048}--\blpage{1064}
(\byear{2022})
\end{bchapter}
\endbibitem

\bibitem[\protect\citeauthoryear{Feijo and Moreira}{2023}]{feijo2023improving}
\begin{barticle}
\bauthor{\bsnm{Feijo}, \binits{D.d.V.}},
\bauthor{\bsnm{Moreira}, \binits{V.P.}}:
\batitle{Improving abstractive summarization of legal rulings through textual entailment}.
\bjtitle{Artificial intelligence and law}
\bvolume{31}(\bissue{1}),
\bfpage{91}--\blpage{113}
(\byear{2023})
\end{barticle}
\endbibitem

\bibitem[\protect\citeauthoryear{Beltagy et~al.}{2020}]{beltagy2020longformer}
\begin{botherref}
\oauthor{\bsnm{Beltagy}, \binits{I.}},
\oauthor{\bsnm{Peters}, \binits{M.E.}},
\oauthor{\bsnm{Cohan}, \binits{A.}}:
Longformer: The long-document transformer.
arXiv preprint arXiv:2004.05150
(2020)
\end{botherref}
\endbibitem

\bibitem[\protect\citeauthoryear{Zhang et~al.}{2020}]{zhang2020pegasus}
\begin{bchapter}
\bauthor{\bsnm{Zhang}, \binits{J.}},
\bauthor{\bsnm{Zhao}, \binits{Y.}},
\bauthor{\bsnm{Saleh}, \binits{M.}},
\bauthor{\bsnm{Liu}, \binits{P.}}:
\bctitle{Pegasus: Pre-training with extracted gap-sentences for abstractive summarization}.
In: \bbtitle{International Conference on Machine Learning},
pp. \bfpage{11328}--\blpage{11339}
(\byear{2020}).
\bcomment{PMLR}
\end{bchapter}
\endbibitem

\bibitem[\protect\citeauthoryear{Teubner et~al.}{2023}]{teubner2023welcome}
\begin{barticle}
\bauthor{\bsnm{Teubner}, \binits{T.}},
\bauthor{\bsnm{Flath}, \binits{C.M.}},
\bauthor{\bsnm{Weinhardt}, \binits{C.}},
\bauthor{\bsnm{Aalst}, \binits{W.}},
\bauthor{\bsnm{Hinz}, \binits{O.}}:
\batitle{Welcome to the era of chatgpt et al. the prospects of large language models}.
\bjtitle{Business \& Information Systems Engineering}
\bvolume{65}(\bissue{2}),
\bfpage{95}--\blpage{101}
(\byear{2023})
\end{barticle}
\endbibitem

\bibitem[\protect\citeauthoryear{Zhang et~al.}{2023}]{zhang2023benchmarking}
\begin{botherref}
\oauthor{\bsnm{Zhang}, \binits{T.}},
\oauthor{\bsnm{Ladhak}, \binits{F.}},
\oauthor{\bsnm{Durmus}, \binits{E.}},
\oauthor{\bsnm{Liang}, \binits{P.}},
\oauthor{\bsnm{McKeown}, \binits{K.}},
\oauthor{\bsnm{Hashimoto}, \binits{T.B.}}:
Benchmarking large language models for news summarization.
arXiv preprint arXiv:2301.13848
(2023)
\end{botherref}
\endbibitem

\bibitem[\protect\citeauthoryear{Lin}{2004}]{lin2004rouge}
\begin{bchapter}
\bauthor{\bsnm{Lin}, \binits{C.-Y.}}:
\bctitle{{ROUGE}: A package for automatic evaluation of summaries}.
In: \bbtitle{Text Summarization Branches Out},
pp. \bfpage{74}--\blpage{81}
(\byear{2004})
\end{bchapter}
\endbibitem

\bibitem[\protect\citeauthoryear{Banerjee and Lavie}{2005}]{banerjee2005meteor}
\begin{bchapter}
\bauthor{\bsnm{Banerjee}, \binits{S.}},
\bauthor{\bsnm{Lavie}, \binits{A.}}:
\bctitle{Meteor: An automatic metric for mt evaluation with improved correlation with human judgments}.
In: \bbtitle{{Proceedings of the ACL Workshop on Intrinsic and Extrinsic Evaluation Measures for Machine Translation And/or Summarization}},
pp. \bfpage{65}--\blpage{72}
(\byear{2005})
\end{bchapter}
\endbibitem

\bibitem[\protect\citeauthoryear{Zhang et~al.}{2019}]{zhang2019bertscore}
\begin{botherref}
\oauthor{\bsnm{Zhang}, \binits{T.}},
\oauthor{\bsnm{Kishore}, \binits{V.}},
\oauthor{\bsnm{Wu}, \binits{F.}},
\oauthor{\bsnm{Weinberger}, \binits{K.Q.}},
\oauthor{\bsnm{Artzi}, \binits{Y.}}:
Bertscore: Evaluating text generation with bert.
arXiv preprint arXiv:1904.09675
(2019)
\end{botherref}
\endbibitem

\bibitem[\protect\citeauthoryear{Laban et~al.}{2022}]{laban2022summac}
\begin{barticle}
\bauthor{\bsnm{Laban}, \binits{P.}},
\bauthor{\bsnm{Schnabel}, \binits{T.}},
\bauthor{\bsnm{Bennett}, \binits{P.N.}},
\bauthor{\bsnm{Hearst}, \binits{M.A.}}:
\batitle{{SummaC: Re-visiting NLI-based models for inconsistency detection in summarization}}.
\bjtitle{Transactions of the Association for Computational Linguistics}
\bvolume{10},
\bfpage{163}--\blpage{177}
(\byear{2022})
\end{barticle}
\endbibitem

\bibitem[\protect\citeauthoryear{Deroy et~al.}{2023}]{deroy2023ready}
\begin{bchapter}
\bauthor{\bsnm{Deroy}, \binits{A.}},
\bauthor{\bsnm{Ghosh}, \binits{K.}},
\bauthor{\bsnm{Ghosh}, \binits{S.}}:
\bctitle{{How Ready are Pre-trained Abstractive Models and LLMs for Legal Case Judgement Summarization?}}
In: \bbtitle{Proceedings of the International Workshop on Artificial Intelligence and Intelligent Assistance for Legal Professionals in the Digital Workplace (LegalAIIA)}
(\byear{2023})
\end{bchapter}
\endbibitem

\bibitem[\protect\citeauthoryear{Ji et~al.}{2023}]{10.1145/3571730}
\begin{botherref}
\oauthor{\bsnm{Ji}, \binits{Z.}},
\oauthor{\bsnm{Lee}, \binits{N.}},
\oauthor{\bsnm{Frieske}, \binits{R.}},
\oauthor{\bsnm{Yu}, \binits{T.}},
\oauthor{\bsnm{Su}, \binits{D.}},
\oauthor{\bsnm{Xu}, \binits{Y.}},
\oauthor{\bsnm{Ishii}, \binits{E.}},
\oauthor{\bsnm{Bang}, \binits{Y.J.}},
\oauthor{\bsnm{Madotto}, \binits{A.}},
\oauthor{\bsnm{Fung}, \binits{P.}}:
Survey of hallucination in natural language generation.
ACM Comput. Surv.
\textbf{55}(12)
(2023)
\end{botherref}
\endbibitem

\bibitem[\protect\citeauthoryear{Bhattacharya et~al.}{2019}]{inbook11}
\begin{bchapter}
\bauthor{\bsnm{Bhattacharya}, \binits{P.}},
\bauthor{\bsnm{Hiware}, \binits{K.}},
\bauthor{\bsnm{Rajgaria}, \binits{S.}},
\bauthor{\bsnm{Pochhi}, \binits{N.}},
\bauthor{\bsnm{Ghosh}, \binits{K.}},
\bauthor{\bsnm{Ghosh}, \binits{S.}}:
\bctitle{A comparative study of summarization algorithms applied to legal case judgments}.
In: \bbtitle{Proc. European Conference on Information Retrieval (ECIR)},
pp. \bfpage{413}--\blpage{428}
(\byear{2019})
\end{bchapter}
\endbibitem

\bibitem[\protect\citeauthoryear{Nigam et~al.}{2023}]{nigam2023nonet}
\begin{bchapter}
\bauthor{\bsnm{Nigam}, \binits{S.K.}},
\bauthor{\bsnm{Deroy}, \binits{A.}},
\bauthor{\bsnm{Shallum}, \binits{N.}},
\bauthor{\bsnm{Mishra}, \binits{A.K.}},
\bauthor{\bsnm{Roy}, \binits{A.}},
\bauthor{\bsnm{Mishra}, \binits{S.K.}},
\bauthor{\bsnm{Bhattacharya}, \binits{A.}},
\bauthor{\bsnm{Ghosh}, \binits{S.}},
\bauthor{\bsnm{Ghosh}, \binits{K.}}:
\bctitle{Nonet at semeval-2023 task 6: Methodologies for legal evaluation}.
In: \bbtitle{Proceedings of the 17th International Workshop on Semantic Evaluation (SemEval-2023)},
pp. \bfpage{1293}--\blpage{1303}
(\byear{2023})
\end{bchapter}
\endbibitem

\bibitem[\protect\citeauthoryear{Nigam and Deroy}{2023}]{nigam2023fact}
\begin{botherref}
\oauthor{\bsnm{Nigam}, \binits{S.K.}},
\oauthor{\bsnm{Deroy}, \binits{A.}}:
Fact-based court judgment prediction.
arXiv preprint arXiv:2311.13350
(2023)
\end{botherref}
\endbibitem

\bibitem[\protect\citeauthoryear{Saravanan et~al.}{2006}]{saravanan2006improving}
\begin{barticle}
\bauthor{\bsnm{Saravanan}, \binits{M.}},
\bauthor{\bsnm{Ravindran}, \binits{B.}},
\bauthor{\bsnm{Raman}, \binits{S.}}:
\batitle{Improving legal document summarization using graphical models}.
\bjtitle{Frontiers in Artificial Intelligence and Applications}
\bvolume{152},
\bfpage{51}
(\byear{2006})
\end{barticle}
\endbibitem

\bibitem[\protect\citeauthoryear{Ahmed and Devanbu}{2023}]{10.1145/3551349.3559555}
\begin{bchapter}
\bauthor{\bsnm{Ahmed}, \binits{T.}},
\bauthor{\bsnm{Devanbu}, \binits{P.}}:
\bctitle{Few-shot training llms for project-specific code-summarization}.
In: \bbtitle{Proceedings of the 37th IEEE/ACM International Conference on Automated Software Engineering}
(\byear{2023})
\end{bchapter}
\endbibitem

\bibitem[\protect\citeauthoryear{Chang et~al.}{2023}]{chang2023booookscore}
\begin{botherref}
\oauthor{\bsnm{Chang}, \binits{Y.}},
\oauthor{\bsnm{Lo}, \binits{K.}},
\oauthor{\bsnm{Goyal}, \binits{T.}},
\oauthor{\bsnm{Iyyer}, \binits{M.}}:
Booookscore: A systematic exploration of book-length summarization in the era of llms.
arXiv preprint arXiv:2310.00785
(2023)
\end{botherref}
\endbibitem

\bibitem[\protect\citeauthoryear{Dimarco}{2023}]{dimarco2023llm}
\begin{botherref}
\oauthor{\bsnm{Dimarco}, \binits{M.H.}}:
Llm-based comment summarization and topic matching for videos
(2023)
\end{botherref}
\endbibitem

\bibitem[\protect\citeauthoryear{Schneider and Turchi}{2023}]{schneider2023team}
\begin{bchapter}
\bauthor{\bsnm{Schneider}, \binits{F.}},
\bauthor{\bsnm{Turchi}, \binits{M.}}:
\bctitle{Team zoom@ automin 2023: Utilizing topic segmentation and llm data augmentation for long-form meeting summarization}.
In: \bbtitle{Proceedings of the 16th International Natural Language Generation Conference: Generation Challenges},
pp. \bfpage{101}--\blpage{107}
(\byear{2023})
\end{bchapter}
\endbibitem

\bibitem[\protect\citeauthoryear{Maity et~al.}{2024}]{maity2024novel}
\begin{bchapter}
\bauthor{\bsnm{Maity}, \binits{S.}},
\bauthor{\bsnm{Deroy}, \binits{A.}},
\bauthor{\bsnm{Sarkar}, \binits{S.}}:
\bctitle{A novel multi-stage prompting approach for language agnostic mcq generation using gpt}.
In: \bbtitle{European Conference on Information Retrieval},
pp. \bfpage{268}--\blpage{277}
(\byear{2024})
\end{bchapter}
\endbibitem

\bibitem[\protect\citeauthoryear{Maity et~al.}{2023}]{maity2023harnessing}
\begin{botherref}
\oauthor{\bsnm{Maity}, \binits{S.}},
\oauthor{\bsnm{Deroy}, \binits{A.}},
\oauthor{\bsnm{Sarkar}, \binits{S.}}:
Harnessing the power of prompt-based techniques for generating school-level questions using large language models.
FIRE 2023,
30
(2023)
\end{botherref}
\endbibitem

\bibitem[\protect\citeauthoryear{Maity et~al.}{2024a}]{maity2024exploring}
\begin{botherref}
\oauthor{\bsnm{Maity}, \binits{S.}},
\oauthor{\bsnm{Deroy}, \binits{A.}},
\oauthor{\bsnm{Sarkar}, \binits{S.}}:
Exploring the capabilities of prompted large language models in educational and assessment applications.
arXiv preprint arXiv:2405.11579
(2024)
\end{botherref}
\endbibitem

\bibitem[\protect\citeauthoryear{Maity et~al.}{2024b}]{maity2024effective}
\begin{botherref}
\oauthor{\bsnm{Maity}, \binits{S.}},
\oauthor{\bsnm{Deroy}, \binits{A.}},
\oauthor{\bsnm{Sarkar}, \binits{S.}}:
How effective is gpt-4 turbo in generating school-level questions from textbooks based on bloom’s revised taxonomy?
(2024)
\end{botherref}
\endbibitem

\bibitem[\protect\citeauthoryear{Deroy and Maity}{2021}]{deroy2021multi}
\begin{botherref}
\oauthor{\bsnm{Deroy}, \binits{A.}},
\oauthor{\bsnm{Maity}, \binits{S.}}:
Multi-label classification of covid-tweets using large language models
(2021)
\end{botherref}
\endbibitem

\bibitem[\protect\citeauthoryear{Maity et~al.}{2024}]{maity2024ready}
\begin{botherref}
\oauthor{\bsnm{Maity}, \binits{S.}},
\oauthor{\bsnm{Deroy}, \binits{A.}},
\oauthor{\bsnm{Sarkar}, \binits{S.}}:
How ready are generative pre-trained large language models for explaining bengali grammatical errors?
arXiv preprint arXiv:2406.00039
(2024)
\end{botherref}
\endbibitem

\bibitem[\protect\citeauthoryear{Deroy and Maity}{2023}]{deroy2023questioning}
\begin{botherref}
\oauthor{\bsnm{Deroy}, \binits{A.}},
\oauthor{\bsnm{Maity}, \binits{S.}}:
Questioning biases in case judgment summaries: Legal datasets or large language models?
arXiv preprint arXiv:2312.00554
(2023)
\end{botherref}
\endbibitem

\bibitem[\protect\citeauthoryear{Ji et~al.}{2023}]{hallucination-survey}
\begin{botherref}
\oauthor{\bsnm{Ji}, \binits{Z.}},
\oauthor{\bsnm{Lee}, \binits{N.}},
\oauthor{\bsnm{Frieske}, \binits{R.}},
\oauthor{\bsnm{Yu}, \binits{T.}},
\oauthor{\bsnm{Su}, \binits{D.}},
\oauthor{\bsnm{Xu}, \binits{Y.}},
\oauthor{\bsnm{Ishii}, \binits{E.}},
\oauthor{\bsnm{Bang}, \binits{Y.J.}},
\oauthor{\bsnm{Madotto}, \binits{A.}},
\oauthor{\bsnm{Fung}, \binits{P.}}:
Survey of hallucination in natural language generation.
ACM Computing Surveys
\textbf{55}(12)
(2023)
\end{botherref}
\endbibitem

\bibitem[\protect\citeauthoryear{Ahmad et~al.}{2023}]{ahmad2023creating}
\begin{botherref}
\oauthor{\bsnm{Ahmad}, \binits{M.}},
\oauthor{\bsnm{Yaramic}, \binits{I.}},
\oauthor{\bsnm{Roy}, \binits{T.D.}}:
Creating trustworthy llms: Dealing with hallucinations in healthcare ai
(2023)
\end{botherref}
\endbibitem

\bibitem[\protect\citeauthoryear{Zhang et~al.}{2023}]{zhang-etal-2023-extractive}
\begin{bchapter}
\bauthor{\bsnm{Zhang}, \binits{S.}},
\bauthor{\bsnm{Wan}, \binits{D.}},
\bauthor{\bsnm{Bansal}, \binits{M.}}:
\bctitle{Extractive is not faithful: An investigation of broad unfaithfulness problems in extractive summarization}.
In: \bbtitle{Proceedings of the 61st Annual Meeting of the Association for Computational Linguistics (Volume 1: Long Papers)},
pp. \bfpage{2153}--\blpage{2174}
(\byear{2023})
\end{bchapter}
\endbibitem

\bibitem[\protect\citeauthoryear{Moro and Ragazzi}{2022}]{moro2022semantic}
\begin{bchapter}
\bauthor{\bsnm{Moro}, \binits{G.}},
\bauthor{\bsnm{Ragazzi}, \binits{L.}}:
\bctitle{Semantic self-segmentation for abstractive summarization of long documents in low-resource regimes}.
In: \bbtitle{Proceedings of the AAAI Conference on Artificial Intelligence},
vol. \bseriesno{36},
pp. \bfpage{11085}--\blpage{11093}
(\byear{2022})
\end{bchapter}
\endbibitem

\bibitem[\protect\citeauthoryear{Zhang et~al.}{2022}]{zhang-etal-2022-summn}
\begin{bchapter}
\bauthor{\bsnm{Zhang}, \binits{Y.}},
\bauthor{\bsnm{Ni}, \binits{A.}},
\bauthor{\bsnm{Mao}, \binits{Z.}},
\bauthor{\bsnm{Wu}, \binits{C.H.}},
\bauthor{\bsnm{Zhu}, \binits{C.}},
\bauthor{\bsnm{Deb}, \binits{B.}},
\bauthor{\bsnm{Awadallah}, \binits{A.}},
\bauthor{\bsnm{Radev}, \binits{D.}},
\bauthor{\bsnm{Zhang}, \binits{R.}}:
\bctitle{{S}umm$^n$: A multi-stage summarization framework for long input dialogues and documents}.
In: \bbtitle{Proceedings of the 60th Annual Meeting of the Association for Computational Linguistics (Volume 1: Long Papers)},
pp. \bfpage{1592}--\blpage{1604}
(\byear{2022})
\end{bchapter}
\endbibitem

\bibitem[\protect\citeauthoryear{Moro and Ragazzi}{2023}]{MORO2023126356}
\begin{barticle}
\bauthor{\bsnm{Moro}, \binits{G.}},
\bauthor{\bsnm{Ragazzi}, \binits{L.}}:
\batitle{Align-then-abstract representation learning for low-resource summarization}.
\bjtitle{Neurocomputing}
\bvolume{548},
\bfpage{126356}
(\byear{2023})
\end{barticle}
\endbibitem

\bibitem[\protect\citeauthoryear{Moro et~al.}{2023}]{moro2023efficient}
\begin{barticle}
\bauthor{\bsnm{Moro}, \binits{G.}},
\bauthor{\bsnm{Ragazzi}, \binits{L.}},
\bauthor{\bsnm{Valgimigli}, \binits{L.}},
\bauthor{\bsnm{Frisoni}, \binits{G.}},
\bauthor{\bsnm{Sartori}, \binits{C.}},
\bauthor{\bsnm{Marfia}, \binits{G.}}:
\batitle{Efficient memory-enhanced transformer for long-document summarization in low-resource regimes}.
\bjtitle{Sensors}
\bvolume{23}(\bissue{7}),
\bfpage{3542}
(\byear{2023})
\end{barticle}
\endbibitem

\bibitem[\protect\citeauthoryear{Zheng and Lapata}{2019}]{zheng-lapata-2019-sentence}
\begin{bchapter}
\bauthor{\bsnm{Zheng}, \binits{H.}},
\bauthor{\bsnm{Lapata}, \binits{M.}}:
\bctitle{Sentence centrality revisited for unsupervised summarization}.
In: \beditor{\bsnm{Korhonen}, \binits{A.}},
\beditor{\bsnm{Traum}, \binits{D.}},
\beditor{\bsnm{M{\`a}rquez}, \binits{L.}} (eds.)
\bbtitle{Proceedings of the 57th Annual Meeting of the Association for Computational Linguistics},
pp. \bfpage{6236}--\blpage{6247}
(\byear{2019})
\end{bchapter}
\endbibitem

\bibitem[\protect\citeauthoryear{Dong et~al.}{2021}]{dong2021discourse}
\begin{bchapter}
\bauthor{\bsnm{Dong}, \binits{Y.}},
\bauthor{\bsnm{Mircea}, \binits{A.}},
\bauthor{\bsnm{Cheung}, \binits{J.C.K.}}:
\bctitle{Discourse-aware unsupervised summarization for long scientific documents}.
In: \bbtitle{Proceedings of the 16th Conference of the European Chapter of the Association for Computational Linguistics: Main Volume},
pp. \bfpage{1089}--\blpage{1102}
(\byear{2021})
\end{bchapter}
\endbibitem

\bibitem[\protect\citeauthoryear{Grusky et~al.}{2018}]{grusky-etal-2018-newsroom}
\begin{bchapter}
\bauthor{\bsnm{Grusky}, \binits{M.}},
\bauthor{\bsnm{Naaman}, \binits{M.}},
\bauthor{\bsnm{Artzi}, \binits{Y.}}:
\bctitle{{N}ewsroom: A dataset of 1.3 million summaries with diverse extractive strategies}.
In: \bbtitle{Proceedings of the 2018 Conference of the North {A}merican Chapter of the Association for Computational Linguistics: Human Language Technologies, Volume 1 (Long Papers)},
pp. \bfpage{708}--\blpage{719}
(\byear{2018})
\end{bchapter}
\endbibitem

\bibitem[\protect\citeauthoryear{Shen et~al.}{2022}]{shen2022multi}
\begin{barticle}
\bauthor{\bsnm{Shen}, \binits{Z.}},
\bauthor{\bsnm{Lo}, \binits{K.}},
\bauthor{\bsnm{Yu}, \binits{L.}},
\bauthor{\bsnm{Dahlberg}, \binits{N.}},
\bauthor{\bsnm{Schlanger}, \binits{M.}},
\bauthor{\bsnm{Downey}, \binits{D.}}:
\batitle{Multi-lexsum: Real-world summaries of civil rights lawsuits at multiple granularities}.
\bjtitle{Advances in Neural Information Processing Systems}
\bvolume{35},
\bfpage{13158}--\blpage{13173}
(\byear{2022})
\end{barticle}
\endbibitem

\bibitem[\protect\citeauthoryear{Liu}{2019}]{liu2019fine}
\begin{botherref}
\oauthor{\bsnm{Liu}, \binits{Y.}}:
Fine-tune bert for extractive summarization.
arXiv preprint arXiv:1903.10318
(2019)
\end{botherref}
\endbibitem

\bibitem[\protect\citeauthoryear{Nallapati et~al.}{2017}]{nallapati2017summarunner}
\begin{bchapter}
\bauthor{\bsnm{Nallapati}, \binits{R.}},
\bauthor{\bsnm{Zhai}, \binits{F.}},
\bauthor{\bsnm{Zhou}, \binits{B.}}:
\bctitle{Summarunner: A recurrent neural network based sequence model for extractive summarization of documents}.
In: \bbtitle{Proceedings of the AAAI Conference on Artificial Intelligence},
vol. \bseriesno{31},
pp. \bfpage{3075}--\blpage{3081}
(\byear{2017})
\end{bchapter}
\endbibitem

\bibitem[\protect\citeauthoryear{Deroy et~al.}{2021}]{deroy2021analytical}
\begin{bchapter}
\bauthor{\bsnm{Deroy}, \binits{A.}},
\bauthor{\bsnm{Bhattacharya}, \binits{P.}},
\bauthor{\bsnm{Ghosh}, \binits{K.}},
\bauthor{\bsnm{Ghosh}, \binits{S.}}:
\bctitle{An analytical study of algorithmic and expert summaries of legal cases}.
In: \bbtitle{Prof. International Conference on Legal Knowledge and Information Systems (JURIX)},
pp. \bfpage{90}--\blpage{99}
(\byear{2021})
\end{bchapter}
\endbibitem

\bibitem[\protect\citeauthoryear{Moro et~al.}{2023}]{moro2023graph}
\begin{bchapter}
\bauthor{\bsnm{Moro}, \binits{G.}},
\bauthor{\bsnm{Ragazzi}, \binits{L.}},
\bauthor{\bsnm{Valgimigli}, \binits{L.}}, \betal:
\bctitle{Graph-based abstractive summarization of extracted essential knowledge for low-resource scenarios}.
In: \bbtitle{26th European Conference on Artificial Intelligence},
vol. \bseriesno{372},
pp. \bfpage{1747}--\blpage{1754}
(\byear{2023})
\end{bchapter}
\endbibitem

\bibitem[\protect\citeauthoryear{Huang et~al.}{2023}]{LLM-hallucination-survey}
\begin{botherref}
\oauthor{\bsnm{Huang}, \binits{L.}},
\oauthor{\bsnm{Yu}, \binits{W.}},
\oauthor{\bsnm{Ma}, \binits{W.}},
\oauthor{\bsnm{Zhong}, \binits{W.}},
\oauthor{\bsnm{Feng}, \binits{Z.}},
\oauthor{\bsnm{Wang}, \binits{H.}},
\oauthor{\bsnm{Chen}, \binits{Q.}},
\oauthor{\bsnm{Peng}, \binits{W.}},
\oauthor{\bsnm{Feng}, \binits{X.}},
\oauthor{\bsnm{Qin}, \binits{B.}},
\oauthor{\bsnm{Liu}, \binits{T.}}:
A Survey on Hallucination in Large Language Models: Principles, Taxonomy, Challenges, and Open Questions
(2023)
\end{botherref}
\endbibitem

\bibitem[\protect\citeauthoryear{Huang et~al.}{2021}]{huang-etal-2021-efficient}
\begin{bchapter}
\bauthor{\bsnm{Huang}, \binits{L.}},
\bauthor{\bsnm{Cao}, \binits{S.}},
\bauthor{\bsnm{Parulian}, \binits{N.}},
\bauthor{\bsnm{Ji}, \binits{H.}},
\bauthor{\bsnm{Wang}, \binits{L.}}:
\bctitle{Efficient attentions for long document summarization}.
In: \bbtitle{Proceedings of the 2021 Conference of the North American Chapter of the Association for Computational Linguistics: Human Language Technologies},
pp. \bfpage{1419}--\blpage{1436}
(\byear{2021})
\end{bchapter}
\endbibitem

\bibitem[\protect\citeauthoryear{Landis and Koch}{1977}]{landis1977measurement}
\begin{botherref}
\oauthor{\bsnm{Landis}, \binits{J.R.}},
\oauthor{\bsnm{Koch}, \binits{G.G.}}:
The measurement of observer agreement for categorical data.
biometrics,
159--174
(1977)
\end{botherref}
\endbibitem

\end{thebibliography}

\end{document}